\documentclass{article} 
\usepackage[preprint]{icml2026}
\usepackage{lineno}
\usepackage{times}  
\usepackage{helvet}  
\usepackage{courier}  
\usepackage{url}  
\usepackage{graphicx} 
\usepackage{natbib}  
\usepackage{caption} 
\usepackage{newfloat}
\usepackage{listings}
\DeclareCaptionStyle{ruled}{labelfont=normalfont,labelsep=colon,strut=off} 
\usepackage{booktabs}

\usepackage[caption=false]{subfig}
\usepackage{verbatim}
\usepackage{amsmath}
\usepackage{amssymb}
\usepackage{amsfonts}
\usepackage{nicefrac}
\usepackage{pgfplots}
    \usepgfplotslibrary{colorbrewer}
\usepackage{colortbl}
\usepackage{microtype}
\usepackage{xcolor}
\usepackage{tikz}
\usepackage{pgfplots}
\usetikzlibrary{arrows.meta}
\usetikzlibrary{calc}

\usepackage{xr-hyper}
\newcommand{\mb}[1]{\mathbf{#1}}

\icmltitlerunning{Dense Local Dependency}

\begin{document}

\twocolumn[
  \icmltitle{Dense Local Dependencies Induce Attention-Logit Explosion 
  and Training Instability During Long-Sequence Transformer Training}



  \icmlsetsymbol{equal}{*}

  \begin{icmlauthorlist}
    \icmlauthor{Suvadeep Hajra}{equal,yyy}
  \end{icmlauthorlist}

  \icmlaffiliation{yyy}{Department of Electrical Engineering, Indian Institute of Technology Delhi, New Delhi, India}

  \icmlcorrespondingauthor{Suvadeep Hajra}{suvadeep.hajra@gmail.com}

  \icmlkeywords{LLM, Pretraining, Training Instability}

  \vskip 0.3in
]



\printAffiliationsAndNotice{}  


\begin{abstract}

Autoregressive transformer language models frequently exhibit 
training instability when trained on long sequences, particularly 
under low-precision arithmetic. Although this
instability is often accompanied by attention-logit explosion, 
its underlying cause remains poorly understood. In this work, 
we present analytical insights and empirical evidence that dense local
dependencies are a major contributor to attention-logit 
explosion. We demonstrate that dense local dependency patterns 
yield an effectively high-rank attention structure, which the
low-rank parameterization of self-attention approximates 
with increasingly large logits as the sequence length grows. This
logit inflation is associated with training instability under
low-precision arithmetic. We support this explanation through 
extensive experiments on synthetic and language modeling tasks. 
Our results consistently show that attention-logit growth 
increases with sequence length, is mitigated by increasing the
attention dimension, and is substantially reduced by explicitly
modeling dense local dependencies. Furthermore, we show that 
this growth is driven by the density of local dependencies 
rather than by locality alone. 
Overall, our findings suggest that explicitly modeling 
dense local dependencies constitutes an important design 
principle for developing stable, efficient, and scalable 
long-context transformer architectures for autoregressive 
language modeling.
\end{abstract}

\section{Introduction}
\label{sec:introduction}
Transformer language models have become the backbone of 
modern machine learning systems, achieving remarkable 
success across diverse domains such as natural language 
processing \cite{VaswaniSPUJGKP17,DevlinCLT19,radford2018improving,radford2019language}, 
computer vision \cite{ChenRC0JLS20,YuXKLBWVKYAHHPLZBW22,abs-2503-17074,ChangZJLF22}, and 
speech \cite{BaevskiZMA20,HsuBTLSM21,AoWZ0RW0KLZWQ0W22,GulatiQCPZYHWZW20}. 
These models have enabled state-of-the-art results in 
applications such as machine translation, document 
summarization, code generation, image captioning, and 
multimodal reasoning.

Despite their immense success, transformer language models
often exhibit training instability, particularly
during large-scale pretraining or when processing 
long sequences 
\cite{abs-2304-09871,ChowdheryNDBMRBCSGSSTMRBTSPRDHPBAI23,LiZH22,WortsmanLXEAACG24,ZhaiLLBR0GS23,0001DMPHGSCGAJB23,NishidaNS24,abs-2502-11034,KediaZKJGL24}. 
This instability manifests as sudden spikes or 
divergence in the training loss after an apparently stable
optimization trajectory. Such failures are particularly 
costly at scale: they can waste substantial computational 
resources, require rollbacks to earlier checkpoints, and, 
in severe cases, restart training entirely. 
As models and context lengths continue to grow, understanding 
and addressing long-sequence training instability becomes 
increasingly important for the reliable development of 
large-scale language and speech models.

Several explanations and solutions have been proposed. 
For instance, \citet{LiuLGCH20} attribute instability 
to amplification of parameter perturbations through 
residual connections. Others have implicated the Adam 
optimizer \citep{KingmaB14} and its interaction with 
large-scale training \citep{abs-2304-09871}. Long-sequence
training itself has been identified as a contributing 
factor, motivating techniques such as progressively 
increasing sequence length during training 
\citep{LiZH22,abs-2108-06084}. Several recent studies 
have observed that instability is often accompanied by
rapid growth of attention logits and have proposed 
normalization-based remedies such as QK-normalization 
\citep{HenryDPC20,WortsmanLXEAACG24,ZhaiLLBR0GS23,0001DMPHGSCGAJB23,KediaZKJGL24}. 
Other approaches include parameter reparameterization 
\citep{NishidaNS24}, specialized optimizers 
\citep{abs-2502-11034}, learning-rate warmup, and 
scaling-rule modifications such as $\mu$Param 
\citep{abs-2203-03466}. 
While these methods often improve stability, they 
primarily address its symptoms or contributing factors.
It remains unclear why longer sequences lead to 
attention-logit growth and training instability, and
what underlying property of long-context modeling drives
this behavior.

\begin{figure*}[t!]
    \centering
    \subfloat[Attention pattern for (causally) full attention.\label{fig:att_full}]{\begin{tikzpicture}[scale=0.95]
  \def\cellsize{0.4cm}
  \foreach \i in {0,...,5} {
    \foreach \j in {0,...,5} {
      \draw (\j*\cellsize, -\i*\cellsize) rectangle ++(\cellsize, -\cellsize);
    }
  }

  \node at (0.2cm, -0.2cm) {\color{blue}1};
  \node at (0.6cm, -0.2cm) {\color{red}x};
  \node at (1.0cm, -0.2cm) {\color{red}x};
  \node at (1.4cm, -0.2cm) {\color{red}x};
  \node at (1.8cm, -0.2cm) {\color{red}x};
  \node at (2.2cm, -0.2cm) {\color{red}x};

  \node at (0.2cm, -0.6cm) {\color{blue}.2};
  \node at (0.6cm, -0.6cm) {\color{blue}.8};
  \node at (1.0cm, -0.6cm) {\color{red}x};
  \node at (1.4cm, -0.6cm) {\color{red}x};
  \node at (1.8cm, -0.6cm) {\color{red}x};
  \node at (2.2cm, -0.6cm) {\color{red}x};

  \node at (0.2cm, -1.0cm) {\color{blue}0};
  \node at (0.6cm, -1.0cm) {\color{blue}.7};
  \node at (1.0cm, -1.0cm) {\color{blue}.3};
  \node at (1.4cm, -1.0cm) {\color{red}x};
  \node at (1.8cm, -1.0cm) {\color{red}x};
  \node at (2.2cm, -1.0cm) {\color{red}x};

  \node at (0.2cm, -1.4cm) {\color{blue}0};
  \node at (0.6cm, -1.4cm) {\color{blue}0};
  \node at (1.0cm, -1.4cm) {\color{blue}.5};
  \node at (1.4cm, -1.4cm) {\color{blue}.5};
  \node at (1.8cm, -1.4cm) {\color{red}x};
  \node at (2.2cm, -1.4cm) {\color{red}x};

  \node at (0.2cm, -1.8cm) {\color{blue}0};
  \node at (0.6cm, -1.8cm) {\color{blue}0};
  \node at (1.0cm, -1.8cm) {\color{blue}0};
  \node at (1.4cm, -1.8cm) {\color{blue}.1};
  \node at (1.8cm, -1.8cm) {\color{blue}.9};
  \node at (2.2cm, -1.8cm) {\color{red}x};

  \node at (0.2cm, -2.2cm) {\color{blue}0};
  \node at (0.6cm, -2.2cm) {\color{blue}0};
  \node at (1.0cm, -2.2cm) {\color{blue}0};
  \node at (1.4cm, -2.2cm) {\color{blue}0};
  \node at (1.8cm, -2.2cm) {\color{blue}.4};
  \node at (2.2cm, -2.2cm) {\color{blue}.6};

  \node at (1.2cm, -2.7cm) {{\bf \scriptsize key}};
  \node[rotate=90] at (-.4cm, -1.2cm) {{\bf \scriptsize query}};

\end{tikzpicture}}\hfill
    \subfloat[Attention pattern for local attention.\label{fig:att_local}]{\begin{tikzpicture}[scale=0.95]
  \def\cellsize{0.4cm}
  \foreach \i in {0,...,5} {
    \foreach \j in {0,...,5} {
      \draw (\j*\cellsize, -\i*\cellsize) rectangle ++(\cellsize, -\cellsize);
    }
  }

  \node at (0.2cm, -0.2cm) {\color{blue}1};
  \node at (0.6cm, -0.2cm) {\color{red}x};
  \node at (1.0cm, -0.2cm) {\color{red}x};
  \node at (1.4cm, -0.2cm) {\color{red}x};
  \node at (1.8cm, -0.2cm) {\color{red}x};
  \node at (2.2cm, -0.2cm) {\color{red}x};

  \node at (0.2cm, -0.6cm) {\color{blue}.2};
  \node at (0.6cm, -0.6cm) {\color{blue}.8};
  \node at (1.0cm, -0.6cm) {\color{red}x};
  \node at (1.4cm, -0.6cm) {\color{red}x};
  \node at (1.8cm, -0.6cm) {\color{red}x};
  \node at (2.2cm, -0.6cm) {\color{red}x};

  \node at (0.2cm, -1.0cm) {\color{red}x};
  \node at (0.6cm, -1.0cm) {\color{blue}.7};
  \node at (1.0cm, -1.0cm) {\color{blue}.3};
  \node at (1.4cm, -1.0cm) {\color{red}x};
  \node at (1.8cm, -1.0cm) {\color{red}x};
  \node at (2.2cm, -1.0cm) {\color{red}x};

  \node at (0.2cm, -1.4cm) {\color{red}x};
  \node at (0.6cm, -1.4cm) {\color{red}x};
  \node at (1.0cm, -1.4cm) {\color{blue}.5};
  \node at (1.4cm, -1.4cm) {\color{blue}.5};
  \node at (1.8cm, -1.4cm) {\color{red}x};
  \node at (2.2cm, -1.4cm) {\color{red}x};

  \node at (0.2cm, -1.8cm) {\color{red}x};
  \node at (0.6cm, -1.8cm) {\color{red}x};
  \node at (1.0cm, -1.8cm) {\color{red}x};
  \node at (1.4cm, -1.8cm) {\color{blue}.1};
  \node at (1.8cm, -1.8cm) {\color{blue}.9};
  \node at (2.2cm, -1.8cm) {\color{red}x};

  \node at (0.2cm, -2.2cm) {\color{red}x};
  \node at (0.6cm, -2.2cm) {\color{red}x};
  \node at (1.0cm, -2.2cm) {\color{red}x};
  \node at (1.4cm, -2.2cm) {\color{red}x};
  \node at (1.8cm, -2.2cm) {\color{blue}.4};
  \node at (2.2cm, -2.2cm) {\color{blue}.6};

  \node at (1.2cm, -2.7cm) {{\bf \scriptsize key}};
  \node[rotate=90] at (-.4cm, -1.2cm) {{\bf \scriptsize query}};

\end{tikzpicture}}\hfill
    \subfloat[Maximum logit vs. sequence length on syn. experiment.\label{fig:syn_logit_growth_comp}]{\begin{tikzpicture}[scale=0.67]
\begin{axis}[
  legend style = {at = {(0.9, 0.95)}, fill=none, fill opacity=0.9, draw opacity=1,text opacity=1, rounded corners=3pt, draw=none, font=\large, legend columns=2},
  legend cell align={left},
  xlabel=\textbf{seq. len $n$},
  ylabel=\textbf{max $|$logit$|$},
  xlabel style={yshift=3pt},
  xmin=100,xmax=2500,
  ymin=0.0,
  ymax=35,
  xmode=log,
  grid,
  grid style={dashed},
  width=.22\textwidth,
  height=.17\textwidth,
  cycle list name = color list,
  xtick = {100, 500, 2500},
  xticklabels = {100, 500, 2500},
  scale only axis = true,
  enlargelimits = false,
  label style={font=\large},
  tick label style={font=\small}  
  ]

\addplot table {figs/data/synthetic_results_new/logitGrowth_softmax_varingLen_qdim25_sparsity0.5_tstep100K.dat};
\addlegendentry{full}
\addplot table {figs/data/synthetic_results_new/logitGrowth_softmaxLocal_varingLen_qdim25_sparsity0.5_tstep100K.dat};
\addlegendentry{local}
\end{axis}
\end{tikzpicture}}\hfill
    \subfloat[Maximum logit vs. local loss on syn. experiment.\label{fig:syn_logit_growth_varingD}]{\begin{tikzpicture}[scale=0.67]
\begin{axis}[
  legend style = {at = {(0.83, 0.97)}, fill=none, fill opacity=0.9, draw opacity=1,text opacity=1, rounded corners=3pt, draw=none, font=\large,legend columns=1},
  legend cell align={left},
  xlabel=\textbf{local loss},
  ylabel=\textbf{max $|$logit$|$},
  xlabel style={yshift=3pt},
  xmin=0.0001,
  ymin=0,
  ymax=45,
  xmode=log,
  grid,
  grid style={dashed},
  width=.23\textwidth,
  height=.17\textwidth,
  cycle list name = color list,
  x dir=reverse, 
  scale only axis = true,
  enlargelimits = false,
  label style={font=\large},
  tick label style={font=\small}  
  ]

\addplot table {figs/data/synthetic_results_new/logit_vs_local_loss_softmax_slen2500_qdim25.dat};
\addlegendentry{full: $d=25$}
\addplot table {figs/data/synthetic_results_new/logit_vs_local_loss_softmax_slen2500_qdim50.dat};
\addlegendentry{full: $d=50$}
\addplot [black!20!brown] table {figs/data/synthetic_results_new/logit_vs_local_loss_softmaxLocal_slen2500_qdim25.dat};
\addlegendentry{local: $d=25$}
\end{axis}
\end{tikzpicture}}
    \caption{    The low-rank bottleneck of full 
    self-attention in representing
    dense local dependencies. (a) Full causal attention
    approximates a dense local attention pattern using the 
    low-rank factorization \(QK^\top\), whose rank is at most 
    \(d\). (b) Sliding-window local attention directly models 
    the dependency band, whose width is independent of sequence
    length. (c) On a synthetic dense-local-dependency task,
    the maximum logit attained by full attention at a fixed
    reconstruction loss over the dependency band (local loss,
    here \(1.46 \times 10^{-4}\)) grows rapidly with sequence
    length \(n\) at fixed attention dimension \(d=25\), whereas
    the local-attention baseline exhibits substantially smaller
    logit magnitudes. (d) Increasing the attention dimension 
    from \(d=25\) to \(d=50\) substantially reduces the observed
    logit growth, although the logits remain larger than those 
    of local attention.}
    \label{fig:local_vs_full_attn}
\end{figure*}

\paragraph{Dense Local Dependencies as a Source of Logit Explosion.}
In this work, we identify dense local dependencies as an important
and previously overlooked contributor to long-sequence training 
instability. Language and speech modeling 
tasks contain substantial local structure, with token predictions 
significantly influenced by nearby context. We show that, when such
local dependencies are dense, the resulting attention patterns
become increasingly difficult for standard full self-attention to
represent as the sequence length grows.

To illustrate this intuition, consider a sequence of input tokens
$\mathrm{X}=[\mathbf{x}_0,\ldots,\mathbf{x}_{n-1}]^T \in 
\mathbb{R}^{n\times d}$. The self-attention operation produces
$\mathrm{Y} = \mathrm{P}\mathrm{X}\mathrm{W}_V$,
where $\mathrm{P}\in\mathbb{R}^{n\times n}$ is the attention matrix
and $\mathrm{W}_V$ is a learnable projection matrix. The attention
matrix is computed as $\mathrm{P}=\mathrm{softmax}(\mathrm{S})
=\mathrm{softmax}(\mathrm{Q}\mathrm{K}^T) =\mathrm{softmax}
\left(\mathrm{X}\mathrm{W}_Q\mathrm{W}_K^T\mathrm{X}^T\right),$
where $\mathrm{Q}$ and $\mathrm{K}$ denote the query and key matrices, 
and $\mathrm{S}$ contains the attention logits.
Although the attention matrix $\mathrm{P}$ can be of full rank, 
its logit matrix $\mathrm{S}=\mathrm{Q}\mathrm{K}^T$ is 
constrained to have rank at most $d$. Consequently, when $n \gg d$, 
self-attention is subject to the \textbf{low-rank bottleneck} 
identified by \citet{BhojanapalliYRR20}. This bottleneck is less
problematic when attention is concentrated on a relatively small set
of important tokens, yielding attention patterns that are effectively
low-rank. In contrast, dense local dependencies induce approximately
banded attention patterns whose rank grows linearly with the sequence
length (Figure~\ref{fig:att_full}). As a result, the gap between
the rank of the target attention pattern and the maximum rank 
attainable by the logit matrix increases 
with sequence length. 
This increasing representational mismatch suggests that
approximating these increasingly complex attention patterns
at a fixed attention dimension may become more challenging
as sequence length grows, potentially contributing to the
observed logit growth.
We propose that this mechanism, rooted in the low-rank 
bottleneck of self-attention, explains the attention-logit 
explosion observed during long-sequence training and, under 
low-precision arithmetic, the training instability.

\paragraph{Empirical Evidence.}
To validate the proposed mechanism, we first construct a synthetic 
learning task in which the target attention matrix contains dense
local dependencies but no meaningful long-range structure
(Section~\ref{sec:limitation_sa}). The 
results show that, at a fixed attention dimension, 
the logit magnitude attained by the trained model at
a given reconstruction accuracy increases rapidly 
with sequence length, whereas 
increasing the attention dimension substantially reduces it 
(Figures~\ref{fig:syn_logit_growth_comp} and
\ref{fig:syn_logit_growth_varingD}). In contrast, the
sliding-window local-attention baseline exhibits consistently 
lower logit growth across sequence lengths. We then conduct 
extensive experiments on transformer training for autoregressive
language modeling (Section~\ref{sec:exp_results}). The results
consistently support the proposed
mechanism: attention-logit growth increases with sequence length,
decreases with larger attention dimensions, and is substantially 
reduced by explicitly modeling dense local dependencies using 
sliding-window local attention heads. Increasing local attention
capacity, either by adding local heads or widening the local 
attention span, further suppresses logit growth in the remaining
full-attention heads, while reducing the density of local 
interactions likewise mitigates logit explosion. Across all 
settings, rapid attention-logit growth precedes 
the onset of training instability.

\paragraph{Implications for Attention Design.}
The above analysis suggests that attention-logit explosion in language
modeling is fundamentally tied to the need for full self-attention to
represent dense local dependencies. Consequently, explicitly modeling
dense local dependencies, either through dedicated local attention
heads (Figure~\ref{fig:att_local}) or locality-biased
positional encodings, provides a natural way
to alleviate attention-logit explosion. Consistent with this
perspective, our empirical results show that replacing a subset of full
attention heads with local attention heads for modeling dense
local dependencies substantially suppresses attention-logit growth and
stabilizes long-sequence training. Moreover, because a local attention
head requires only $O(nw)$ computation, where the local attention span
$w \ll n$, compared to the $O(n^2)$ complexity of a full attention
head, this decomposition also improves the computational efficiency of
long-context transformer models. It additionally reduces the KV-cache
requirements during autoregressive inference.

Overall, our findings establish dense local dependencies as 
a key source of attention-logit explosion and motivate 
locality-aware attention as a principled direction for stable 
and efficient long-context transformer architectures.

\section{Related Work}

\paragraph{Low-Rank Bottleneck of Self-Attention}
Bhojanapalli et al.~\cite{BhojanapalliYRR20} identified a low-rank
bottleneck in self-attention, showing it cannot represent all
attention matrices $\mathrm{P}$ when the attention dimension is
smaller than the sequence length $n$. Their remedy, setting the
head dimension to $n$, is impractical for large $n$, raising the
attention cost to $O(n^3)$. We identify learning dense local
dependencies as a practically important setting where this 
bottleneck is especially severe. Our analysis and experiments
show that this bottleneck drives attention-logit growth and 
long-sequence training instability.

\paragraph{Expressivity of Self-Attention}
Likhosherstov et al.~\cite{LikhosherstovCW23} showed that 
sufficiently sparse attention matrices, including 
bounded-width banded patterns arising from dense 
local dependencies, can be approximated with an 
embedding dimension of only $d=O(\log n)$. However, 
this guarantee is weak: zero entries are approximated
by small positive values, and the resulting approximation
error at the attention output may grow linearly
with $n$. In contrast, we study how attention-logit 
magnitude varies with attention dimension and sequence
length during transformer training, as optimization 
progressively reduces approximation error. We find 
that trained models exhibit increasingly large logits
when representing high-rank patterns induced by dense
local dependencies as $n$ grows, contributing to 
attention-logit explosion and, under low-precision 
arithmetic, training instability.

\section{Dense Local Dependencies and the Low-Rank Attention Bottleneck}
\label{sec:limitation_sa}
In this section, we analyze how dense local dependency patterns
interact with the low-rank bottleneck of full self-attention.
We show that, at a fixed attention-head dimension, representing
such patterns becomes increasingly challenging as sequence
length grows, with the trained models exhibiting progressively
larger attention logits.
We then validate this mechanism through a controlled
synthetic task, using sliding-window local attention as a
reference baseline. To formalize this analysis, we consider an
autoregressive task over sequences of length $n$, where the
prediction of the next token depends only on the immediately
preceding $l$ tokens, with $l \ll n$. For this task, the ideal
attention matrix $\mathrm{P}\in\mathbb{R}^{n\times n}$ satisfies
$\mathrm{P}[q,k]>0$ for $0<q-k\le l$ and
$\mathrm{P}[q,k]=0$ otherwise.

When attempting to learn this dependency pattern 
using full causal attention, the model aims 
to approximate a matrix $\mathrm{P}'$ using
the self-attention operation such that 
$\mathrm{P}'[q, k] = \mathrm{P}[q, k]$ for 
$q - k \ge 0$, and treats $\mathrm{P}'[q, k]$ as
a ``don't care'' term for $q - k < 0$ (since these 
terms are masked in causal attention). An 
illustration of such an attention pattern is 
shown in Figure~\ref{fig:att_full}, where $n=6$ 
and $l=2$; red entries represent masked (don’t care) 
terms. Importantly, $\mathrm{P}'$ is a matrix of 
rank $n$, which grows linearly with the sequence 
length. During training, the attention
mechanism attempts to replicate $\mathrm{P}'$ using
$\mathrm{softmax}((\mathrm{Q}\mathrm{K}^T+\mathrm{M}_{\mathcal{S}})/\sqrt{d})$,
where $\mathrm{Q}=[\mb{q}_0,\ldots,\mb{q}_{n-1}]^T\in \mathbb{R}^{n\times d}$,
$\mathrm{K}=[\mb{k}_0, \ldots, \mb{k}_{n-1}]^T \in \mathbb{R}^{n\times d}$
are the query and key matrices and 
$\mathrm{M}_{\mathcal{S}}\in \mathbb{R}^{n\times n}$ be 
the causal mask (i.e., $\mathrm{M}_{\mathcal{S}}[q, k]=0$ for
$q-k\ge 0$ and $-\infty$ otherwise). 
Since \(QK^\top\) has rank at most \(d\), the pre-softmax
logits are subject to a low-rank constraint. Although 
the row-wise softmax can produce full-rank attention 
matrices, representing dense local attention patterns
with a low-dimensional logit parameterization can 
become increasingly difficult as the sequence length
grows. We hypothesize that this representational 
difficulty contributes to the logit growth observed 
during training and investigate this hypothesis 
empirically in the following sections.

\subsection{Validation through a Synthetic Task}
We now construct a controlled synthetic task of learning
dense local dependencies to directly test
this prediction. We use sliding-window local attention
as a baseline. Because the local 
baseline only models interactions within a window of width 
$w \approx l \ll n$, the effective rank of the target pattern it must 
represent is bounded by the window size rather than the 
sequence length (Figure~\ref{fig:att_local}). 
If the proposed low-rank bottleneck contributes to logit 
growth, full attention should exhibit progressively larger
logits than local attention as the sequence length increases.

\paragraph{Synthetic Task}
We consider the task of learning the relation 
$\mathrm{Y}=\mathrm{PV}$ by a self-attention operation
where $\mathrm{V}=[\mb{v}_0,\ldots,\mb{v}_{n-1}]^T \in \mathbb{R}^{n\times d_v}$
is the value matrix and 
$\mathrm{Y}=[\mb{y}_0,\ldots,\mb{y}_{n-1}]^T \in \mathbb{R}^{n\times d_v}$
is the corresponding output. The target attention matrix $\mathrm{P}\in\mathbb{R}^{n\times n}$ is fixed, and the 
model learns query and key matrices such that 
$\mathrm{P} \approx \mathrm{softmax}((\mathrm{Q}\mathrm{K}^T+\mathrm{M}_{\mathcal{S}})/\sqrt{d})$
where $\mathrm{Q},\mathrm{K}\in\mathbb{R}^{n\times d}$ are the
learnable query and key matrices. 


To model dense local dependencies, $\mathbf{P}$ is generated as 
a causal banded matrix with nonzero entries only for $0<q-k\le50$.
Each nonzero entry is sampled independently from a Bernoulli$(0.5)$
distribution, and each row is normalized to obtain a valid attention 
distribution. To isolate the effect of the attention mechanism itself,
we set $\mathrm{V}=\mathrm{I}$, the identity matrix, so that 
$\mathrm{Y}=\mathrm{P}$ and the optimization objective becomes  
reconstruction of the target attention matrix.

\paragraph{Local-Attention Baseline}
The baseline replaces the full causal mask with a sliding-window causal
mask of span $w=50$, matching the width of the ground-truth dependency
band. In other words, we replace the 
full causal mask $\mathrm{M}_S$ with a sliding-window 
causal mask $\mathrm{M}_S^{w}$ that permits each query to
attend only to $w=50$ past keys: $M_S^w[q,k]=0$ for $0 < q - k < 50$
and $-\infty$ otherwise.

\paragraph{Training Setup}
We train both full and local self-attention for 100K steps using the
Adam optimizer and MSE loss. However, because the global MSE becomes
increasingly dominated by the quadratically growing number of zero
entries as sequence length increases, even a nearly 
constant attention distribution can achieve small 
global MSE for large $n$. We therefore report the
{\em local loss}, computed only over the dependency band
$\{(q,k):0<q-k\le50\}$.

\paragraph{Results}
Figure~\ref{fig:syn_logit_growth_comp} reports the maximum
absolute attention logit attained by full attention when
achieving a local loss of \(1.46 \times 10^{-4}\),
corresponding to the minimum local loss attained after
$100K$ training steps. We compare sequence lengths 
\(n = 100, 500,\) and \(2500\), while fixing the attention
dimension at \(d = 25\). The figure confirms the predicted
trend: at a fixed attention dimension, the maximum logit
observed at this reconstruction accuracy grows rapidly
with sequence length, whereas the reference local-attention 
baseline exhibits substantially smaller logit magnitudes
and remains nearly constant between \(n = 500\) and 
\(n = 2500\). Figure~\ref{fig:syn_logit_growth_varingD} 
further shows that increasing the attention dimension from
\(d = 25\) to \(d = 50\), while keeping \(n = 2500\), 
substantially reduces the observed logit magnitude, 
although it remains larger than that of the 
local-attention baseline.

\section{Empirical Validation on Language Data}
\label{sec:exp_results}
We evaluate whether the findings from the synthetic task extend
to language modeling. Using GPT-2 Small on the PG-19 dataset,
we first examine how sequence length and attention-head dimension
affect attention-logit growth and training stability, thereby
validating the role of the low-rank bottleneck. We then investigate
the role of dense local dependencies by varying the local attention
capacity and the density of local interactions in an attention
architecture that combines full and local attention heads.
Additional results on a second language dataset and a speech
dataset are provided in Appendix~\ref{app:other_datasets}.

\subsection{Experimental Setup}
\label{sec:exp_setup}

\paragraph{Model Architecture}
For all experiments, we use an autoregressive decoder-only
GPT-2 architecture as the base architecture. Unless stated 
otherwise, we use GPT-2 Small with $12$ transformer layers 
($L$), a hidden dimension of $768$ ($d_{model}$), $12$ 
attention heads ($H$), and a feedforward dimension of $3072$. 
The multi-head self-attention (MHSA) module is implemented
using FlashAttention~\citep{dao2023flashattention2}.


\paragraph{Training Details}
We trained all models using the AdamW optimizer 
with a weight decay of $1\text{e}{-1}$, 
$\beta_1 = 0.9$, and $\beta_2 = 0.95$. Gradient
clipping was applied with a maximum norm of 
$1.0$. The learning rate followed a cosine 
decay schedule with linear warmup: the maximum 
learning rate was set to $6\text{e}{-4}$, the
minimum learning rate was set to $6\text{e}{-5}$, with $2K$ warmup
steps and a total of $600K$ decay steps.
Across all experiments, we fixed the total 
number of tokens per batch to $2^{19}$. 
Consequently, when using longer sequence 
lengths, we proportionally reduced the number
of sequences per batch to maintain a constant
token budget. Unless stated otherwise, we used 
mixed-precision training with the bfloat16
(BF16) data type. However, all reported 
attention-logit measurements were obtained 
from separate FP32 training runs. We found that,
in unstable BF16 runs, numerical errors can 
produce logit spikes that obscure 
the underlying growth behavior. The FP32 runs 
therefore provide a more faithful illustration
of the intrinsic logit dynamics of the model.

\paragraph{Dataset and Preprocessing}
We conducted our experiments on the PG-19 dataset 
\cite{RaePJHL20}, a collection of English-language 
books. All texts were normalized using 
NMT\_NFKC and tokenized with a SentencePiece unigram
model with a vocabulary size of $10K$.
We use PG-19 because its long documents make it well
suited for studying long-sequence autoregressive 
language modeling. 


\begin{figure}[t!]
    \centering
    \subfloat[Logit growth for sequence lengths $n=512$, $2K$, and $8K$ with head dimension $d=64$.\label{fig:logit_explosion_longSeq}]{\begin{tikzpicture}[scale=1]
\begin{axis}[
  legend style = {at = {(0.96, 0.99)}, fill=none, fill opacity=0.9, draw opacity=1,text opacity=1, rounded corners=3pt, draw=none, font=\scriptsize,
  legend columns=2},
  legend cell align={left},
  legend image code/.code={
    \draw[#1] (0cm,0cm) -- (0.5cm,0cm);
  },
  xlabel=\textbf{training step},
  ylabel=\textbf{max $|$logit$|$},
  ylabel style={yshift=-15pt},
  xlabel style={yshift=8pt},
  x tick scale label style={
    yshift=8pt
  },
  xmin=2000,
  xmax=14000,
  ymin=20,
  ymax=950,
  grid,
  grid style={dashed},
  width=.17\textwidth,
  height=.09\textwidth,
  cycle list name = color list,
  scale only axis = true,
  enlargelimits = false,
  ytick={200, 400, 600, 800},
  yticklabels={200, 400, 600, 800},
  label style={font=\scriptsize},
  tick label style={font=\scriptsize},
  ]

\addplot [thick, black!20!blue] table {figs/data/logit_explotion/softmax_L12_H12_HDim64_blkSize512.dat};
\addlegendentry{n: 512}
\addplot [thick, black!20!green] table 
{figs/data/logit_explotion/softmax_L12_H12_HDim64_blkSize2048_fp32.dat};
\addlegendentry{n: 2K}
\addplot [thick, black!20!red] table 
{figs/data/logit_explotion/softmax_L12_H12_HDim64_blkSize8192_fp32.dat};
\addlegendentry{n: 8K}
\end{axis}
\end{tikzpicture}} \hfill
    \subfloat[Training curves for sequence lengths $n=512$, $2K$, and $8K$ with head dimension $d=64$.\label{fig:instability_longSeq}]{\begin{tikzpicture}[scale=1]
\begin{axis}[
  legend style = {at = {(0.98, 0.98)}, fill=none, fill opacity=0.9, draw opacity=1,text opacity=1, rounded corners=3pt, draw=none, font=\scriptsize},
  legend columns=2,
  legend cell align={left},
  legend image code/.code={
    \draw[#1] (0cm,0cm) -- (0.5cm,0cm);
  },
  x filter/.code={\pgfmathparse{#1}\pgfmathresult},
  xlabel=\textbf{training step},
  ylabel=\textbf{train loss},
  ylabel style={yshift=-18pt},
  xlabel style={yshift=8pt},
  x tick scale label style={
    yshift=8pt
  },
  xmin=0,
  xmax=25000,
  ymin=2.5,
  ymax=9.5,
  grid,
  grid style={dashed},
  width=.17\textwidth,
  height=.09\textwidth,
  cycle list name = color list,
  scale only axis = true,
  enlargelimits = false,
  label style={font=\scriptsize},
  tick label style={font=\scriptsize}  
  ]

\addplot [thick, black!20!blue] table 
{figs/data/fig_instability/softmax_L12_H12_HDim64_blkSize512.dat};
\addlegendentry{n: 512}
\addplot [thick, black!20!green] table 
{figs/data/fig_instability/softmax_L12_H12_HDim64_blkSize2048.dat};
\addlegendentry{n: 2K}
\addplot [thick, black!20!red] table 
{figs/data/fig_instability/softmax_L12_H12_HDim64_blkSize8192.dat};
\addlegendentry{n: 8K}
\end{axis}
\end{tikzpicture}}\\
    \subfloat[Logit growth  for head dimensions $d=64, 128, 256, $ and
    $384$ with $n=2K$.\label{fig:logit_explosion_largeDim}]{\begin{tikzpicture}[scale=1]
\begin{axis}[
  legend style = {at = {(0.94, 0.99)}, fill=none, fill opacity=0.9, draw opacity=1,text opacity=1, rounded corners=3pt, draw=none, font=\scriptsize,
  legend columns=2},
  legend cell align={left},
  legend image code/.code={
    \draw[#1] (0cm,0cm) -- (0.5cm,0cm);
  },
  xlabel=\textbf{training step},
  ylabel=\textbf{max $|$logit$|$},
  ylabel style={yshift=-15pt},
  xlabel style={yshift=8pt},
  x tick scale label style={
    yshift=8pt
  },
  xmin=2000,
  xmax=14000,
  ymin=20,
  ymax=950,
  grid,
  grid style={dashed},
  width=.17\textwidth,
  height=.09\textwidth,
  cycle list name = color list,
  scale only axis = true,
  enlargelimits = false,
  ytick={200, 400, 600},
  yticklabels={200, 400, 600, 800},
  label style={font=\scriptsize},
  tick label style={font=\scriptsize}  
  ]

\addplot [thick, black!20!blue] table {figs/data/logit_explotion/softmax_L12_H12_HDim64_blkSize2048_fp32.dat};
\addlegendentry{d: 64}
\addplot [thick, black!20!green] table 
{figs/data/logit_explotion/softmax_L12_H12_HDim128_blkSize2048_fp32.dat};
\addlegendentry{d: 128}
\addplot [thick, black!20!red] table 
{figs/data/logit_explotion/softmax_L12_H12_HDim256_blkSize2048.dat};
\addlegendentry{d: 256}
\addplot [thick, black!20!cyan] table 
{figs/data/logit_explotion/softmax_L12_H12_HDim384_blkSize2048.dat};
\addlegendentry{d: 384}
\end{axis}
\end{tikzpicture}}\hfill
    \subfloat[Training curves for head dimensions $d=64$, $128$, $256$, and $384$ with $n=2K$.\label{fig:instability_largeDim}]{\begin{tikzpicture}[scale=1]
\begin{axis}[
  legend style = {at = {(0.95, 0.95)}, fill=none, fill opacity=0.9, draw opacity=1,text opacity=1, rounded corners=3pt, draw=none, font=\scriptsize,
  legend columns=2},
  legend cell align={left},
  legend image code/.code={
    \draw[#1] (0cm,0cm) -- (0.5cm,0cm);
  },
  x filter/.code={\pgfmathparse{#1}\pgfmathresult},
  xlabel=\textbf{training step},
  ylabel=\textbf{train loss},
  ylabel style={yshift=-18pt},
  xlabel style={yshift=6.5pt},
  x tick scale label style={
    yshift=8pt
  },
  xmin=0,
  xmax=18000,
  ymin=2.5,
  ymax=9.5,
  grid,
  grid style={dashed},
  width=.17\textwidth,
  height=.09\textwidth,
  cycle list name = color list,
  scale only axis = true,
  enlargelimits = false,
  label style={font=\scriptsize},
  tick label style={font=\scriptsize}  
  ]

\addplot [thick, black!20!blue] table 
{figs/data/fig_instability/softmax_L12_H12_HDim64_blkSize2048.dat};
\addlegendentry{d: 64}
\addplot [thick, black!20!green] table 
{figs/data/fig_instability/softmax_L12_H12_HDim128_blkSize2048.dat};
\addlegendentry{d: 128}
\addplot [thick, black!20!red] table 
{figs/data/fig_instability/softmax_L12_H12_HDim256_blkSize2048.dat};
\addlegendentry{d: 256}
\addplot [thick, black!20!cyan] table 
{figs/data/fig_instability/softmax_L12_H12_HDim384_blkSize2048.dat};
\addlegendentry{d: 384}
\end{axis}
\end{tikzpicture}}
    \caption{Effect of sequence length and head dimension on 
    attention-logit growth and training stability. Increasing
    sequence length amplifies attention-logit growth and 
    instability, whereas increasing the head dimension
    suppresses both.}
    \label{fig:instability_gpt}
\end{figure}

\subsection{Evidence for the Role of Low-Rank Bottleneck 
in Logit Explosion and Training Instability}
\label{sec:low_rank_bottleneck_validation}

The proposed mechanism identifies the low-rank bottleneck as a
primary cause of attention-logit explosion. Consequently, it 
predicts that increasing sequence length should exacerbate 
attention-logit growth, whereas increasing the attention 
dimension should mitigate it. Having confirmed these predictions
in the synthetic task, we now investigate whether they also hold
during language model training. We further show that 
low-precision arithmetic amplifies the resulting logit growth 
into training instability.

\subsubsection{Increasing Sequence Length Worsens Logit
Explosion and Instability}
We examine maximum absolute attention-logit growth during 
training for sequence lengths $n=512$, $2K$, and $8K$,
with a fixed head dimension of $d=64$. As shown in 
Figure~\ref{fig:logit_explosion_longSeq}, logit growth
increases substantially with sequence length. 
Correspondingly, Figure~\ref{fig:instability_longSeq}
shows that BF16 training remains stable at $n=512$ but 
diverges at $n=2K$ and $8K$, with instability closely 
tracking logit growth. These findings extend the 
synthetic results to language modeling, suggesting 
that longer sequences exacerbate the low-rank bottleneck, 
contributing to larger attention logits and greater
training instability.

\subsubsection{Increasing Head Dimension Mitigates Logit 
Explosion and Instability}
We fix the sequence length at n=2K and vary the attention-head dimension
from 64 to 384. As shown in Figure~\ref{fig:logit_explosion_largeDim},
increasing the head dimension substantially suppresses logit
growth: for d=64 and 128, logits continue to grow throughout
training, whereas for d=256 and 384, they plateau and eventually
decrease. Correspondingly, Figure~\ref{fig:instability_largeDim}
shows that models with d=64 and 128 become unstable, while those
with d=256 and 384 remain stable. These results mirror the 
synthetic task, confirming that increasing the head dimension
alleviates the low-rank bottleneck and suppresses attention-logit
growth during language model training.

\subsubsection{Low-Precision Training Amplifies the Instability
Caused by Bottleneck-Induced Logit Growth.}
While the low-rank bottleneck drives attention-logit growth,
low-precision arithmetic determines whether this growth
translates into training instability. To investigate this,
we repeat the MHSA experiment with $d=64$ for $n=2K$ and 
$8K$ using FP32 training. As shown in 
Figure~\ref{fig:instability_precision}, FP32 training remains 
stable in settings where BF16 diverges, indicating that 
low-precision arithmetic is considerably less tolerant of 
large attention logits than FP32. These results show that 
low-precision arithmetic amplifies the effects of 
attention-logit explosion, allowing it to manifest as 
training instability.

Taken together, these results provide consistent empirical 
support for the role of the low-rank bottleneck: 
logit growth scales with sequence length, is mitigated by 
larger head dimensions,
and leads to training instability under low-precision arithmetic. 
Having established the low-rank bottleneck as the mechanism
underlying attention-logit explosion, we now investigate 
dense local dependencies as its primary source.

\begin{figure}
    \centering
    \subfloat[$n = 2K$.\label{fig:instability_fp32_2K}]
    {\begin{tikzpicture}[scale=1]
\begin{axis}[
  legend style = {at = {(0.95, 0.95)}, fill=none, fill opacity=0.9, draw opacity=1,text opacity=1, rounded corners=3pt, draw=none, font=\scriptsize,
  legend columns=2},
  legend cell align={left},
  legend style={
        /tikz/every even column/.append style={column sep=0.1cm}
    },
  legend image code/.code={
    \draw[#1] (0cm,0cm) -- (0.5cm,0cm);
  },
  x filter/.code={\pgfmathparse{#1}\pgfmathresult},
  xlabel=\textbf{training step},
  ylabel=\textbf{train loss},
  ylabel style={yshift=-15pt},
  xlabel style={yshift=8pt},
  x tick scale label style={
    yshift=8pt
  },
  xmin=0,
  xmax=15000,
  ymin=2.5,
  ymax=9.5,
  grid,
  grid style={dashed},
  width=.17\textwidth,
  height=.09\textwidth,
  cycle list name = color list,
  xtick = {5000, 10000},
  scale only axis = true,
  enlargelimits = false,
  label style={font=\scriptsize},
  tick label style={font=\scriptsize}  
  ]

\addplot [thick, black!20!blue] table 
{figs/data/fig_instability/softmax_L12_H12_HDim64_blkSize2048.dat};
\addlegendentry{BF16}
\addplot [thick, black!20!red] table 
{figs/data/fig_instability/softmax_L12_H12_HDim64_blkSize2048_fp32.dat};
\addlegendentry{FP32}
\end{axis}
\end{tikzpicture}}\hfill
    \subfloat[$n = 8K$.\label{fig:instability_fp32_8K}]
    {\begin{tikzpicture}[scale=1]
\begin{axis}[
  legend style = {at = {(0.95, 0.95)}, fill=none, fill opacity=0.9, draw opacity=1,text opacity=1, rounded corners=3pt, draw=none, font=\scriptsize,
  legend columns=2},
  legend cell align={left},
  legend style={
        /tikz/every even column/.append style={column sep=0.1cm}
    },
  legend image code/.code={
    \draw[#1] (0cm,0cm) -- (0.5cm,0cm);
  },
  x filter/.code={\pgfmathparse{#1}\pgfmathresult},
  xlabel=\textbf{training step},
  ylabel=\textbf{train loss},
  ylabel style={yshift=-15pt},
  xlabel style={yshift=8pt},
  x tick scale label style={
    yshift=8pt
  },
  xmin=0,
  xmax=15000,
  ymin=2.5,
  ymax=9.5,
  grid,
  grid style={dashed},
  width=.17\textwidth,
  height=.09\textwidth,
  cycle list name = color list,
  xtick = {5000, 10000},
  scale only axis = true,
  enlargelimits = false,
  label style={font=\scriptsize},
  tick label style={font=\scriptsize}  
  ]

\addplot [thick, black!20!blue] table 
{figs/data/fig_instability/softmax_L12_H12_HDim64_blkSize8192.dat};
\addlegendentry{BF16}
\addplot [thick, black!20!red] table 
{figs/data/fig_instability/softmax_L12_H12_HDim64_blkSize8192_fp32.dat};
\addlegendentry{FP32}
\end{axis}
\end{tikzpicture}}
    \caption{Distinction between logit explosion and training 
    instability. Logit explosion leads to training instability 
    primarily in low-precision settings such as BF16 training, 
    whereas FP32 training remains substantially more stable.}
    \label{fig:instability_precision}
\end{figure}

\subsection{Validating Dense Local Dependencies as the Source 
of the Low-Rank Bottleneck}
\label{sec:dense_local_dependency_validation}
The controlled synthetic experiments in 
Section~\ref{sec:limitation_sa} showed that representing
dense local dependencies requires progressively larger 
attention logits, mirroring the growth observed during 
long-sequence transformer training. We now investigate
whether dense local dependencies are a primary source 
of this bottleneck in language model training. To this
end, we employ \textbf{Full-Local Attention}, which 
partitions MHSA heads into two groups: sliding-window 
local-attention heads that model dense local dependencies
and full-attention heads that capture long-range 
dependencies. If dense local dependencies drive the 
bottleneck, this decomposition should reduce logit
growth and improve training stability.

\begin{figure}[t!]
    \centering
    \subfloat[Logit Growth in Full-Local Attention with $H_s=1$ and $H_l=11$.\label{fig:logit_explosion_ls_1_11}]
    {\begin{tikzpicture}[scale=1]
\begin{axis}[
  legend style = {at = {(0.96, 1.05)}, fill=none, fill opacity=0.9, draw opacity=1,text opacity=1, rounded corners=3pt, draw=none, font=\scriptsize,
  legend columns=1},
  legend cell align={left},
  legend image code/.code={
    \draw[#1] (0cm,0cm) -- (0.5cm,0cm);
  },
  xlabel=\textbf{training step},
  ylabel=\textbf{max $|$logit$|$},
  ylabel style={yshift=-15pt},
  xlabel style={yshift=5pt},
  x tick scale label style={
    yshift=5pt
  },
  xmin=2000,
  xmax=14000,
  ymin=20,
  ymax=1000,
  grid,
  grid style={dashed},
  width=.17\textwidth,
  height=.09\textwidth,
  cycle list name = color list,
  scale only axis = true,
  enlargelimits = false,
  ytick={200, 400, 600, 800},
  label style={font=\scriptsize},
  tick label style={font=\scriptsize}  
  ]

\addplot [thick, black!20!red] table 
{figs/data/logit_explotion/softmax_L12_H12_HDim64_blkSize8192_fp32.dat};
\addlegendentry{MHSA; n: 8K}
\addplot [thick, black!20!cyan] table {figs/data/logit_explotion/softmaxLocalAttn_L12_H12_G11_HDim64_blkSize2048.dat};
\addlegendentry{Full-Local; n: 2K}
\addplot [thick, black!20!violet] table {figs/data/logit_explotion/softmaxLocalAttn_L12_H12_G11_HDim64_blkSize8192.dat};
\addlegendentry{Full-Local; n: 8K}
\addplot [thick, black!20!brown] table {figs/data/logit_explotion/softmaxLocalAttn_L12_H12_G11_HDim64_blkSize32768.dat};
\addlegendentry{Full-Local; n: 32K}
\end{axis}
\end{tikzpicture}} \hfill
    \subfloat[Logit Growth in Full-Local Attention with $H_s=11$ and $H_l=1$.\label{fig:logit_explosion_ls_11_1}]
    {\begin{tikzpicture}[scale=1]
\begin{axis}[
  legend style = {at = {(0.96, 1.05)}, fill=none, fill opacity=0.9, draw opacity=1,text opacity=1, rounded corners=3pt, draw=none, font=\scriptsize,
  legend columns=1},
  legend cell align={left},
  legend image code/.code={
    \draw[#1] (0cm,0cm) -- (0.5cm,0cm);
  },
  xlabel=\textbf{training step},
  ylabel=\textbf{max $|$logit$|$},
  ylabel style={yshift=-15pt},
  xlabel style={yshift=5pt},
  x tick scale label style={
    yshift=5pt
  },
  xmin=2000,
  xmax=14000,
  ymin=20,
  ymax=1000,
  grid,
  grid style={dashed},
  width=.17\textwidth,
  height=.09\textwidth,
  cycle list name = color list,
  scale only axis = true,
  enlargelimits = false,
  ytick={200, 400, 600,800},
  label style={font=\scriptsize},
  tick label style={font=\scriptsize}  
  ]

\addplot [thick, black!20!red] table 
{figs/data/logit_explotion/softmax_L12_H12_HDim64_blkSize8192_fp32.dat};
\addlegendentry{MHSA; n: 8K}
\addplot [thick, black!20!cyan] table {figs/data/logit_explotion/softmaxLocalAttn_L12_H12_HDim64_blkSize2048.dat};
\addlegendentry{Full-Local; n: 2K}
\addplot [thick, black!20!violet] table {figs/data/logit_explotion/softmaxLocalAttn_L12_H12_HDim64_blkSize8192.dat};
\addlegendentry{Full-Local; n: 8K}
\addplot [thick, black!20!brown] table {figs/data/logit_explotion/softmaxLocalAttn_L12_H12_HDim64_blkSize32768.dat};
\addlegendentry{Full-Local; n: 32K}
\end{axis}
\end{tikzpicture}} \\
    \subfloat[Training curves of Full-Local Attention with $H_s=1$ and $H_l=11$.\label{fig:instability_ls_1_11}]
    {\begin{tikzpicture}[scale=1]
\begin{axis}[
  legend style = {at = {(0.95, 0.95)}, fill=none, fill opacity=0.9, draw opacity=1,text opacity=1, rounded corners=3pt, draw=none, font=\scriptsize},
  legend cell align={left},
  legend image code/.code={
    \draw[#1] (0cm,0cm) -- (0.5cm,0cm);
  },
  x filter/.code={\pgfmathparse{#1}\pgfmathresult},
  xlabel=\textbf{training step},
  ylabel=\textbf{train loss},
  ylabel style={yshift=-15pt},
  xlabel style={yshift=5pt},
  x tick scale label style={
    yshift=5pt
  },
  xmin=0,
  xmax=20000,
  ymin=2.5,
  ymax=9.5,
  grid,
  grid style={dashed},
  width=.17\textwidth,
  height=.09\textwidth,
  cycle list name = color list,
  xtick = {5000, 10000, 15000},
  scale only axis = true,
  enlargelimits = false,
  label style={font=\scriptsize},
  tick label style={font=\scriptsize}  
  ]

\addplot [thick, black!20!blue] table 
{figs/data/fig_instability/softmaxLocalAttn_L12_H12_G11_HDim64_attnSpan100_blkSize2048.dat};
\addlegendentry{n: 2K}
\addplot [thick, black!20!green] table 
{figs/data/fig_instability/softmaxLocalAttn_L12_H12_G11_HDim64_attnSpan100_blkSize8192.dat};
\addlegendentry{n: 8K}
\addplot [thick, black!20!red] table 
{figs/data/fig_instability/softmaxLocalAttn_L12_H12_G11_HDim64_attnSpan100_blkSize32768.dat};
\addlegendentry{n: 32K}
\end{axis}
\end{tikzpicture}} \hfill
    \subfloat[Training curves of Full-Local Attention with $H_s=11$ and $H_l=1$.\label{fig:instability_ls_11_1}]
    {\begin{tikzpicture}[scale=1]
\begin{axis}[
  legend style = {at = {(0.95, 0.95)}, fill=none, fill opacity=0.9, draw opacity=1,text opacity=1, rounded corners=3pt, draw=none, font=\scriptsize},
  legend cell align={left},
  legend image code/.code={
    \draw[#1] (0cm,0cm) -- (0.5cm,0cm);
  },
  x filter/.code={\pgfmathparse{#1}\pgfmathresult},
  xlabel=\textbf{training step},
  ylabel=\textbf{train loss},
  ylabel style={yshift=-15pt},
  xlabel style={yshift=5pt},
  x tick scale label style={
    yshift=5pt
  },
  xmin=0,
  xmax=20000,
  ymin=2.5,
  ymax=9.5,
  grid,
  grid style={dashed},
  width=.17\textwidth,
  height=.09\textwidth,
  cycle list name = color list,
  xtick = {5000, 10000, 15000},
  scale only axis = true,
  enlargelimits = false,
  label style={font=\scriptsize},
  tick label style={font=\scriptsize}  
  ]

\addplot [thick, black!20!blue] table 
{figs/data/fig_instability/softmaxLocalAttn_L12_H12_HDim64_attnSpan100_blkSize2048.dat};
\addlegendentry{n: 2K}
\addplot [thick, black!20!green] table 
{figs/data/fig_instability/softmaxLocalAttn_L12_H12_HDim64_attnSpan100_blkSize8192.dat};
\addlegendentry{n: 8K}
\addplot [thick, black!20!red] table 
{figs/data/fig_instability/softmaxLocalAttn_L12_H12_HDim64_attnSpan100_blkSize32768.dat};
\addlegendentry{n: 32K}
\end{axis}
\end{tikzpicture}}
    \caption{Effect of sequence length on attention-logit growth
    and training stability for Full-Local Attention.
    We evaluate two Full-Local Attention configurations: $(H_s,H_l)=(1,11)$ 
    (Figures~\ref{fig:logit_explosion_ls_1_11} and 
    \ref{fig:instability_ls_1_11}) and $(H_s,H_l)=(11,1)$ 
    (Figures~\ref{fig:logit_explosion_ls_11_1} and 
    \ref{fig:instability_ls_11_1}), where $H_s$ and $H_l$ denote 
    the numbers of local and full-attention heads, respectively. 
    The logit-growth plots also include vanilla MHSA 
    ($s=0$, $l=12$) with sequence length $n=8K$ for comparison. 
    Across both Full-Local Attention configurations, logit growth 
    remains substantially lower than MHSA, even for sequence
    lengths up to $32K$, and no training instability is observed
    within the first $20K$ training steps.}
    \label{fig:instability_gpt}
\end{figure}

\subsubsection{Full-Local Attention Reduces Logit Growth and Stabilizes
Training in Long Sequences}

Figures~\ref{fig:logit_explosion_ls_1_11} and
\ref{fig:logit_explosion_ls_11_1} show the maximum attention 
logit for Full-Local Attention under two configurations,
$(H_s,H_l)=(1,11)$ and $(H_s,H_l)=(11,1)$, where $H_s$ and
$H_l$ represent the number of local and full attention heads,
respectively, across sequence 
lengths $n=2K$, $8K$, and $32K$, together with MHSA at $n=8K$ 
for reference. In both configurations, Full-Local Attention exhibits 
substantially smaller logit growth than MHSA, with maximum 
logits at $n=32K$ remaining well below those of MHSA at $n=8K$.
Crucially, because logits are measured across all attention heads
and layers, the reduction is not limited to the local heads: 
allocating dedicated local heads also reduces logit growth in 
the remaining full-attention heads, consistent with our proposed
mechanism that identifies dense local dependencies as a major
contributor to attention-logit growth in MHSA.

Figures~\ref{fig:instability_ls_1_11} and
\ref{fig:instability_ls_11_1} show the corresponding training 
curves. In contrast to MHSA, which becomes unstable beyond 
$n=2K$, Full-Local Attention remains stable at $n=32K$. Notably, 
stable training is achieved even with a single local attention 
head, suggesting that even minimal reallocation of attention 
capacity toward local dependencies substantially mitigates 
logit explosion and training instability.

\begin{figure}
    \centering
    \subfloat[Max. logits in local and full heads in
    Full-Local Attention for $H_s\in \{1,6,11\}$ with $w=100$.\label{fig:local_heads_vs_max_logits}]
    {\begin{tikzpicture}[scale=1]
\begin{axis}[
  legend style = {at = {(0.95, 0.96)}, fill=none, fill opacity=0.9, draw opacity=1,text opacity=1, rounded corners=3pt, draw=none, font=\scriptsize,
  legend columns=1},
  legend cell align={left},
  legend image code/.code={
    \draw[#1] (0cm,0cm) -- (0.45cm,0cm);
  },
  x filter/.code={\pgfmathparse{12-#1}\pgfmathresult},
  xlabel=\textbf{\# of Local Heads $H_s$},
  ylabel=\textbf{max $|$logit$|$},
  ylabel style={yshift=-15pt},
  xlabel style={yshift=5pt},
  x tick scale label style={
    yshift=5pt
  },
  ymin=0,
  ymax=500,
  grid,
  grid style={dashed},
  width=.155\textwidth,
  height=.09\textwidth,
  cycle list name = color list,
  scale only axis = true,
  enlargelimits = false,
  xtick={1, 6, 11},
  ytick={0, 200, 400},
  label style={font=\scriptsize},
  tick label style={font=\scriptsize}  
  ]

\addplot [thick, black!20!blue] table {figs/data/logit_explotion/nFullHeads_vs_localLogits_L12_H12_HDim64_attnSpan100_blkSize32K.dat};
\addlegendentry{Local Heads}
\addplot [thick, black!20!red] table {figs/data/logit_explotion/nFullHeads_vs_fullLogits_L12_H12_HDim64_attnSpan100_blkSize32K.dat};
\addlegendentry{Full Heads}
\end{axis}
\end{tikzpicture}}\hfill
    \subfloat[Max. logits in local and full heads in
    Full-Local Attention for $w\in (10,400)$ with  $H_s=11$.\label{fig:attn_span_vs_max_logits}]
    {\begin{tikzpicture}[scale=1]
\begin{axis}[
  legend style = {at = {(0.95, 0.96)}, fill=none, fill opacity=0.9, draw opacity=1,text opacity=1, rounded corners=3pt, draw=none, font=\scriptsize,
  legend columns=1},
  legend cell align={left},
  legend image code/.code={
    \draw[#1] (0cm,0cm) -- (0.45cm,0cm);
  },
  xlabel=\textbf{Local Attn Span $w$},
  ylabel=\textbf{max $|$logit$|$},
  ylabel style={yshift=-15pt},
  xlabel style={yshift=5pt},
  x tick scale label style={
    yshift=5pt
  },
  ymin=0,
  ymax=500,
  xmode=log,
  grid,
  grid style={dashed},
  width=.155\textwidth,
  height=.09\textwidth,
  cycle list name = color list,
  scale only axis = true,
  enlargelimits = false,
  xtick={10, 20, 50, 100, 200, 400},
  xticklabels={10, 20, 50, 100, 200, 400},
  ytick={0, 200, 400},
  label style={font=\scriptsize},
  tick label style={font=\scriptsize}  
  ]

\addplot [thick, black!20!blue] table {figs/data/logit_explotion/attnSpan_vs_localLogits_L12_H12_G1_HDim64_blkSize32K.dat};
\addlegendentry{Local Heads}
\addplot [thick, black!20!red] table {figs/data/logit_explotion/attnSpan_vs_fullLogits_L12_H12_G1_HDim64_blkSize32K.dat};
\addlegendentry{Full Heads}
\end{axis}
\end{tikzpicture}}
    \caption{Effect of local attention capacity in
    Full-Local Attention on 
    attention-logit growth at $n=32K$.
    (a) Increasing the number of local heads $H_s$ 
    substantially reduces logit growth in the 
    full-attention 
    heads. (b) Reducing the local attention span $w$
    substantially increases the logit growth in the 
    global heads.}
    \label{fig:logit_explosion_varing_params}
\end{figure}

\subsubsection{More Local Attention Capacity Reduces Logit 
Growth in Full Attention Heads}
We investigate how local attention capacity affects 
logit growth in the remaining full attention heads by 
varying (i) the number of local heads $H_s$, keeping 
the total number of heads fixed at $12$ and the span 
fixed at $w=100$, and (ii) the local attention span 
$w$, keeping $H_s=11$ and $H_l=1$, and the sequence 
length fixed at $n=32\mathrm{K}$.

As shown in Figure~\ref{fig:local_heads_vs_max_logits}, 
increasing $H_s$ from $1$ to $11$ substantially reduces 
the logits of the full attention heads, while those of 
the local heads increase only slightly. This indicates 
that allocating more heads to local attention relieves 
the representational burden on the full attention heads.
Figure~\ref{fig:attn_span_vs_max_logits} shows that 
increasing $w$ has a similar effect. 
Together, these results show that increasing local 
attention capacity, either through more local heads 
or a larger span, systematically reduces logit growth 
in the full attention heads, suggesting that large 
logits in MHSA originate from local dependencies.

\subsubsection{High Density of Local Dependencies Causes Logit Explosion}
The previous results suggest that local interactions contribute to
logit growth in full attention, but do not distinguish whether this
behavior arises from locality itself or from the density of local
interactions. Although dense local dependencies can induce
high-rank attention patterns, high rank alone does not necessarily
imply large logit growth (see Section~\ref{subsec:insufficiency_of_high_rank}). 
In sparse high-rank patterns,
removing a small number of connections can separate the pattern into
multiple approximately independent attention blocks. Representing
these blocks separately may require smaller logit magnitudes. 
This suggests that the density of
local interactions, rather than rank alone, may influence logit
growth. We therefore investigate the role of local-interaction
density in the observed behavior.

To test this, we directly manipulate the density of local
interactions by randomly masking a fraction $p$ of attention
connections within a fixed local neighborhood (token pairs $(q,k)$
with $q-k<100$) in MHSA. The masking pattern is fixed across all
heads, layers, and training samples. As shown in
Figure~\ref{fig:localDenseness}, for a small masking fraction
($p=0.1$), attention logits grow rapidly, leading to training
instability. Masking half of the local connections ($p=0.5$)
still results in substantial logit growth during the initial stage
of training, causing unstable optimization. In contrast, masking
$90\%$ of the local connections ($p=0.9$) keeps attention-logit
growth low and stabilizes training.

\begin{figure}[t!]
    \centering
    \subfloat[Logit growth in MHSA under different local 
    dependency densities ($n=8K$).
    \label{fig:logit_explosion_localDenseness}]
    {\begin{tikzpicture}[scale=1]
\begin{axis}[
  legend style = {at = {(0.9, 1.0)}, fill=none, fill opacity=0.6, draw opacity=1,text opacity=1, rounded corners=3pt, draw=none, font=\scriptsize,
  legend columns=2},
  legend cell align={left},
  legend image code/.code={
    \draw[#1] (0cm,0cm) -- (0.5cm,0cm);
  },
  xlabel=\textbf{training step},
  ylabel=\textbf{max $|$logit$|$},
  ylabel style={yshift=-10pt},
  xlabel style={yshift=5pt},
  x tick scale label style={
    yshift=5pt
  },
  xmin=1000,
  xmax=8000,
  ymin=0,
  ymax=2100,
  grid,
  grid style={dashed},
  width=.17\textwidth,
  height=.09\textwidth,
  cycle list name = color list,
  scale only axis = true,
  enlargelimits = false,
  ytick={500, 1000, 1500},
  xtick={3000, 6000},
  label style={font=\scriptsize},
  tick label style={font=\scriptsize}  
  ]

\addplot [thick, black!20!blue] table {figs/data/dense_localAttn_logit_explosion/softmaxKeyDropLocal_L12_H12_HDim64_KDrop0.1_blkSize8K_fp32.dat};
\addlegendentry{p: 0.1}
\addplot [thick, black!20!red] table {figs/data/dense_localAttn_logit_explosion/softmaxKeyDropLocal_L12_H12_HDim64_KDrop0.5_blkSize8K_fp32.dat};
\addlegendentry{p: 0.5}
\addplot [thick, black!20!green] table {figs/data/dense_localAttn_logit_explosion/softmaxKeyDropLocal_L12_H12_HDim64_KDrop0.9_blkSize8K.dat};
\addlegendentry{p: 0.9}
\end{axis}
\end{tikzpicture}}\hfill
    \subfloat[Training curves of MHSA under different 
    local dependency densities ($n=8K$).
    \label{fig:instability_localDenseness}]
    {\begin{tikzpicture}[scale=1]
\begin{axis}[
  legend style = {at = {(0.95, 1.0)}, fill=none, fill opacity=0.9, draw opacity=1,text opacity=1, rounded corners=3pt, draw=none, font=\scriptsize,
  legend columns=2},
  legend cell align={left},
  legend image code/.code={
    \draw[#1] (0cm,0cm) -- (0.5cm,0cm);
  },
  xlabel=\textbf{training step},
  ylabel=\textbf{train loss},
  ylabel style={yshift=-22pt},
  xlabel style={yshift=5pt},
  x tick scale label style={
    yshift=5pt
  },
  xmin=0,
  xmax=12500,
  ymin=3,
  ymax=10,
  grid,
  grid style={dashed},
  width=.18\textwidth,
  height=.09\textwidth,
  cycle list name = color list,
  scale only axis = true,
  enlargelimits = false,
  ytick={4, 6, 8},
  label style={font=\scriptsize},
  tick label style={font=\scriptsize}  
  ]

\addplot [thick, black!20!blue] table 
{figs/data/dense_localAttn_logit_explosion/gpt2_softmaxKeyDropLocal_L12_H12_HDim64_KDrop0.1_blkSize8K.dat};
\addlegendentry{p: 0.1}
\addplot [thick, black!20!red] table {figs/data/dense_localAttn_logit_explosion/gpt2_softmaxKeyDropLocal_L12_H12_HDim64_KDrop0.5_blkSize8K.dat};
\addlegendentry{p: 0.5}
\addplot [thick, black!20!green] table {figs/data/dense_localAttn_logit_explosion/gpt2_softmaxKeyDropLocal_L12_H12_HDim64_KDrop0.9_blkSize8K.dat};
\addlegendentry{p: 0.9}
\end{axis}
\end{tikzpicture}}
    \caption{Effect of varying the density of local 
    dependencies in MHSA. To control the density of local 
    interactions available to the model, we randomly mask 
    a fraction $p$ of attention connections within the
    local neighborhood ($q-k<100$). The masked positions 
    are fixed across all attention heads, layers, and 
    training samples. Larger values of $p$ correspond to 
    sparser local dependency patterns available to the
    model.}
    \label{fig:localDenseness}
\end{figure}

\subsubsection{High Rank Alone Does Not Explain Attention-Logit Growth}
\label{subsec:insufficiency_of_high_rank}

The density ablation suggests that local-interaction density
influences attention-logit growth. To investigate whether this
behavior can be explained by the high rank of such attention 
patterns, we
analyze the relationship between effective rank and logit
magnitude across attention heads, followed by a closer examination
of representative heads with contrasting behaviors.

For each head, we compute the effective rank
of its softmax attention matrix $P$, following \citep{roy2007effective}:
$\operatorname{erank}(P)=\exp(-\sum_i\bar\sigma_i\log\bar\sigma_i)$,
where $\bar\sigma_i$ are the singular values normalized to sum to
one. We also measure the maximum absolute attention logit
$|q^\top k|/\sqrt{d}$ for each head.
Across the 144 heads, effective rank is positively correlated with
the final logit magnitude (Spearman $\rho=0.74$;
Figure~\ref{fig:erank_vs_logit}a). However, effective rank alone
does not explain the variation across heads. We therefore compare
two representative heads: the head with the highest effective rank
(layer~5, head~0) and the head with the largest final logit
(layer~2, head~11). Despite its higher effective rank, the
layer-5 head exhibits substantially lower logit growth than the
layer-2 head (Figure~\ref{fig:erank_vs_logit}b).

Their attention patterns reveal a corresponding structural
difference (Figure~\ref{fig:erank_vs_logit}c). The layer-5 head
concentrates nearly all its attention on the immediately preceding
token ($q-k=1$), yielding a sparse attention pattern. In contrast,
the layer-2 head distributes attention across a wider range of
nearby tokens, forming a denser local pattern.
One possible explanation is that some sparse attention patterns
can be approximated by multiple approximately independent,
lower-rank blocks, each requiring comparatively smaller logit
magnitudes. Dense local patterns, in contrast, may contain
interactions that are more difficult to separate into such blocks.
This comparison suggests that effective rank alone is insufficient
to explain attention-logit growth.

\begin{figure}[t]
  \centering
\begin{tikzpicture}[scale=1]
\begin{axis}[
  name=pa,
  grid, grid style={dashed},
  width=0.085\textwidth, height=.09\textwidth, scale only axis = true,
  label style={font=\scriptsize}, tick label style={font=\scriptsize},
  ylabel near ticks, xlabel near ticks,
  ylabel style={font=\scriptsize, yshift=-2pt}, xlabel style={font=\scriptsize, yshift=2pt},
  xmode=log, ymode=log,
  ytick={1000,10000}, yticklabels={1K,10K},
  xtick={10,100,1000}, xticklabels={10,100,1K},
  log ticks with fixed point,
  xlabel=\textbf{max $|$logit$|$}, ylabel=\textbf{effective rank},
  enlarge x limits=0.05, enlarge y limits=0.08, clip mode=individual,
  title={Spearman $\rho=0.74$}, title style={font=\scriptsize, yshift=-5pt},
  ]
\addplot [only marks, mark=*, mark size=0.6pt, black!90, mark options={fill opacity=0.9, draw opacity=0}]
  table [x=growth, y=erank] {figs/data/fig_erank/heads.dat};
\addplot [only marks, mark=*, mark size=1.3pt, black!20!blue] table [x=growth, y=erank] {figs/data/fig_erank/head_rank.dat};
\addplot [only marks, mark=*, mark size=1.3pt, black!20!red] table [x=growth, y=erank] {figs/data/fig_erank/head_logit.dat};
\end{axis}
\begin{axis}[
  name=pb, at={($(pa.east)+(0.065\textwidth,0)$)}, anchor=west,
  grid, grid style={dashed},
  width=0.085\textwidth, height=.09\textwidth, scale only axis = true,
  label style={font=\scriptsize}, tick label style={font=\scriptsize},
  ylabel near ticks, xlabel near ticks,
  ylabel style={font=\scriptsize, yshift=-2pt}, xlabel style={font=\scriptsize, yshift=2pt},
  ytick={3000,6000}, yticklabels={3K,6K},
  xtick={400,800}, xticklabels={400,800},
  log ticks with fixed point,
  xlabel=\textbf{max $|$logit$|$}, ylabel=\textbf{effective rank},
  enlarge x limits=0.05, enlarge y limits=0.08, clip mode=individual,
  legend to name=headleg, legend columns=1,
  legend style={draw=none, font=\scriptsize, row sep=-2pt}, legend cell align={left},
  legend image code/.code={\draw[#1] (0cm,0cm) -- (0.4cm,0cm);},
  ]
\addplot [thick, black!20!blue, mark=*, mark size=0.5pt] table [x=x, y=erank] {figs/data/fig_erank/traj_rank.dat};
\addlegendentry{max eff.\ rank (L5H0)}
\addplot [thick, black!20!red, mark=*, mark size=0.5pt] table [x=x, y=erank] {figs/data/fig_erank/traj_logit.dat};
\addlegendentry{max logit (L2H11)}
\addplot [only marks, mark=*, mark size=1.3pt, black!20!blue, mark indices={13}, forget plot] table [x=x, y=erank] {figs/data/fig_erank/traj_rank.dat};
\addplot [only marks, mark=*, mark size=1.3pt, black!20!red, mark indices={13}, forget plot] table [x=x, y=erank] {figs/data/fig_erank/traj_logit.dat};
\end{axis}
\begin{axis}[
  name=pc, at={($(pb.east)+(0.065\textwidth,0)$)}, anchor=west,
  grid, grid style={dashed},
  width=0.085\textwidth, height=.09\textwidth, scale only axis = true,
  label style={font=\scriptsize}, tick label style={font=\scriptsize},
  ylabel near ticks, xlabel near ticks,
  ylabel style={font=\scriptsize, yshift=-2pt}, xlabel style={font=\scriptsize, yshift=2pt},
  xmin=0, ymin=0,
  xmax=15, ymax=1,
  xlabel=\textbf{distance $q-k$}, ylabel=\textbf{attn.\ prob.},
  enlargelimits=false,
  ]
\addplot [thick, black!20!blue] table [x=dist, y=prob] {figs/data/fig_erank/dist_rank.dat};
\addplot [thick, black!20!red] table [x=dist, y=prob] {figs/data/fig_erank/dist_logit.dat};
\end{axis}
\node [anchor=south] at ($(pb.north)!0.5!(pc.north)+(0,2pt)$) {\ref{headleg}};
\end{tikzpicture}
  \caption{\textbf{Effective rank alone does not explain logit growth.}
  (a) Logit growth against effective rank of the softmax attention, one point per head
  (Spearman $\rho$ shown).
  (b) Attention logit growth during training for the head with the highest effective rank (blue)
  and the head with the largest final logit (red).
  (c) Mean attention probability against query--key distance for the same two heads.}
  \label{fig:erank_vs_logit}
\end{figure}

Taken together, these results show that attention-logit growth
increases with sequence length and is mitigated by larger head
dimensions, dedicated local attention heads, and sparser local
interactions. Under low-precision arithmetic, this growth can
manifest as training instability. These findings support the
hypothesis that dense local dependencies are an important
contributor to attention-logit growth and long-sequence training
instability. The severity of this effect also appears to increase
with model scale, as discussed in
Section~\ref{sec:larger_model_more_instability}. Beyond stability,
these findings yield useful design principles, discussed in
Section~\ref{sec:attention_design_implications}. Additional results
on two more datasets (language and speech) are provided in
Appendix~\ref{app:other_datasets}.

\section{Dense Local Dependency-Induced Logit Explosion 
Intensifies with Model Scale}
\label{sec:larger_model_more_instability}
Long-sequence training instability also becomes more pronounced 
with model scale. Table~\ref{tab:instability-runs} shows that, 
across both $n = 2\mathrm{K}$ and $n = 4\mathrm{K}$, the number 
of unstable runs (out of four) increases monotonically with 
model size, indicating greater susceptibility to instability
in larger models.


\begin{table}[t]
    \centering
    \begin{footnotesize}
    \begin{tabular}{c|cccc}
    \toprule
     Seq. Len & \multicolumn{4}{c}{Model Size} \\
     \cmidrule(lr){2-5}
     ($n$) & 6.5M & 35.4M & 100.3M & 733.5M \\
    \midrule
     2K & 0 & 0 & 2 & 4 \\
     4K & 0 & 2 & 4 & 4 \\
    \bottomrule
    \end{tabular}
    \end{footnotesize}
    \caption{Number of training runs (out of 4) that become unstable
    within the first 25K training steps, across four model sizes and
    two sequence lengths. The count increases monotonically with model
    size, indicating that larger models are more susceptible to
    long-sequence training instability.}
    \label{tab:instability-runs}
\end{table}

To investigate whether this increased instability is related
to dense local dependencies, we compare GPT-2 Small and GPT-2
Large using Full-Local Attention with one full-attention head
(\(H_l=1\)) and $11$ and $19$ local-attention heads (\(H_s\)),
respectively, at a sequence length of $n=32\mathrm{K}$,
under two local attention spans, $w=10$ and $w=100$. As shown 
in Figure~\ref{fig:larger_models_more_logit_exploration}, GPT-2 
Large exhibits substantially larger logit growth than GPT-2 
Small when the local attention span is small ($w=10$). 
Increasing the span to $w=100$ suppresses the logit growth of 
GPT-2 Large. These results suggest that larger models rely on 
broader local dependency structures and therefore require 
greater local attention capacity to avoid the low-rank 
bottleneck.

Consequently, dense local dependency-induced
attention-logit explosion may become increasingly 
severe as model scale grows, making larger models 
more susceptible to long-sequence training instability
when local dependencies are not adequately modeled. 
This observation may also help explain the loss 
spikes reported in the largest models by
\citet{ChowdheryNDBMRBCSGSSTMRBTSPRDHPBAI23}, although
establishing a causal relationship would require
dedicated investigation.

\section{Implications for Attention Design}
\label{sec:attention_design_implications}

The evidence presented throughout this work suggests that
attention-logit explosion in long-sequence language modeling
is a consequence 
of requiring full self-attention to represent dense local
dependency patterns. This observation points to a broader
design principle: attention mechanisms should devote
dedicated capacity to modeling dense local interactions,
rather than requiring full attention heads to represent both
short- and long-range dependencies.

This perspective helps explain the effectiveness of a broad
class of locality-aware transformer architectures.
Full-Local Attention provides one explicit realization of this
principle, while locality-biased positional encodings such
as RoPE~\citep{SuALPBL24} and ALiBi~\citep{PressSL22}
implicitly encourage attention toward nearby tokens.
Likewise, several structured attention
mechanisms~\cite{abs-1904-10509,abs-2004-05150,ZaheerGDAAOPRWY20,JiangLZWLAHA0L024,GuoQXZ19}
allocate part of their attention capacity to local
interactions. As shown 
in Appendix~\ref{app:additional_evidence},
these approaches
consistently exhibit improved long-sequence training
stability, suggesting that emphasizing dense local
dependencies is a broadly effective way to alleviate the
low-rank bottleneck.
This interpretation also explains why addressing the 
representational bottleneck is more effective than 
suppressing its symptoms. As shown 
in Appendix~\ref{app:alt_train_stab},
methods that explicitly
model or implicitly emphasize dense local dependencies 
outperform existing stabilization techniques, such as 
QK-normalization~\citep{HenryDPC20}, 
Z-loss~\citep{ChowdheryNDBMRBCSGSSTMRBTSPRDHPBAI23}, and 
the AdaGC optimizer~\citep{abs-2502-11034}, which primarily
target large attention logits or optimization stability 
rather than the underlying low-rank bottleneck.

Beyond stability, explicitly modeling dense local dependencies
also brings efficiency benefits. Replacing a subset of full
attention heads with local heads reduces much of the attention
computation from $O(n^2)$ to $O(nw)$, where $w \ll n$ is the
local attention span, while also lowering KV-cache memory during
autoregressive generation. Locality-aware attention can thus
simultaneously improve stability, efficiency, and scalability for
long-context transformers.

\begin{figure}[t!]
    \centering
    \subfloat[Logit growth for $w=10$.
    \label{fig:logit_explosion_gpt2_small}]
    {\begin{tikzpicture}[scale=1]
\begin{axis}[
  legend style = {at = {(0.8, 0.98)}, fill=none, fill opacity=0.9, draw opacity=1,text opacity=1, rounded corners=3pt, draw=none, font=\scriptsize,
  legend columns=1},
  legend cell align={left},
  legend image code/.code={
    \draw[#1] (0cm,0cm) -- (0.45cm,0cm);
  },
  x filter/.code={\pgfmathparse{#1}\pgfmathresult},
  xlabel=\textbf{training step},
  ylabel=\textbf{max. logit},
  ylabel style={yshift=-10pt},
  xlabel style={yshift=5pt},
  x tick scale label style={
    yshift=5pt
  },
  xmin=2000,
  xmax=10000,
  ymin=0,
  ymax=1200,
  grid,
  grid style={dashed},
  width=.155\textwidth,
  height=.09\textwidth,
  cycle list name = color list,
  scale only axis = true,
  enlargelimits = false,
  xtick={4000, 8000},
  ytick={0, 500, 1000},
  scaled y ticks=true,
  label style={font=\scriptsize},
  tick label style={font=\scriptsize}  
  ]

\addplot [thick, black!20!blue] table {figs/data/larger_models_more_logit_explosion/blkSize32768_ckpt_softmaxLocalAttn_L12_H12_G1_HDim64_attnSpan10_blkSize32K.dat};
\addlegendentry{GPT-2 Small}
\addplot [thick, black!20!red] table {figs/data/larger_models_more_logit_explosion/blkSize32768_ckpt_softmaxLocalAttn_L36_H20_G1_HDim64_attnSpan10_blkSize32K_fp32.dat};
\addlegendentry{GPT-2 Large}
\end{axis}
\end{tikzpicture}}\hfill
    \subfloat[Logit growth for $w=100$.
    \label{fig:logit_explosion_gpt2_large}]
    {\begin{tikzpicture}[scale=1]
\begin{axis}[
  legend style = {at = {(0.8, 0.98)}, fill=none, fill opacity=0.9, draw opacity=1,text opacity=1, rounded corners=3pt, draw=none, font=\scriptsize,
  legend columns=1},
  legend cell align={left},
  legend image code/.code={
    \draw[#1] (0cm,0cm) -- (0.45cm,0cm);
  },
  x filter/.code={\pgfmathparse{#1}\pgfmathresult},
  xlabel=\textbf{training step},
  ylabel=\textbf{max. logit},
  ylabel style={yshift=-10pt},
  xlabel style={yshift=5pt},
  x tick scale label style={
    yshift=5pt
  },
  xmin=2000,
  xmax=10000,
  ymin=0,
  ymax=1200,
  grid,
  grid style={dashed},
  width=.16\textwidth,
  height=.09\textwidth,
  cycle list name = color list,
  scale only axis = true,
  enlargelimits = false,
  xtick={4000, 8000},
  ytick={0, 500, 1000},
  label style={font=\scriptsize},
  tick label style={font=\scriptsize}  
  ]

\addplot [thick, black!20!blue] table {figs/data/larger_models_more_logit_explosion/blkSize32768_ckpt_softmaxLocalAttn_L12_H12_G1_HDim64_attnSpan100_blkSize32K.dat};
\addlegendentry{GPT-2 Small}
\addplot [thick, black!20!red] table {figs/data/larger_models_more_logit_explosion/blkSize32768_ckpt_softmaxLocalAttn_L36_H20_G1_HDim64_attnSpan100_blkSize32K.dat};
\addlegendentry{GPT-2 Large}
\end{axis}
\end{tikzpicture}}
    
    \caption{Effect of local attention capacity on logit growth
    for GPT-2 Small and GPT-2 Large using Full-Local Attention
    at $n=32\mathrm{K}$. For a smaller local
    attention capacity ($w=10$), GPT-2 Large exhibits 
    substantially larger logit growth than GPT-2 Small, while 
    logit growth is reduced for both models at $w=100$. This 
    suggests that larger models require greater local attention 
    capacity to suppress logit growth.}
    \label{fig:larger_models_more_logit_exploration}
\end{figure}

\section{Conclusion}
This paper identifies \textbf{dense local dependencies} as a 
major contributor to attention-logit explosion and long-sequence
training instability in autoregressive transformer models. 
We show that dense local dependencies are associated with
increased attention-logit growth during long-sequence 
training. Our experiments suggest that this behavior 
is related to the representational bottleneck imposed
by the low-rank parameterization of self-attention and
contributes to training instability, particularly 
under low-precision arithmetic.
Through analytical insight and extensive empirical validation, we 
demonstrate that this bottleneck becomes more severe as sequence
length increases, is alleviated by increasing the attention-head
dimension, and appears to intensify with model scale. We further
show that explicitly modeling dense local dependencies by 
replacing several full-attention heads with sliding-window local
attention heads substantially suppresses attention-logit growth
and stabilizes long-sequence training. More broadly, our findings 
suggest that explicitly modeling dense local dependencies 
constitutes an important design principle for developing stable, 
efficient, and scalable long-context transformer architectures.

\bibliography{refs}
\bibliographystyle{icml2026}


\appendix

\begin{figure*}[h]
    \centering
    \subfloat[Logit growth in MHSA for increasing $n$ in 
    Wiki40B dataset.\label{fig:logit_explosion_wiki40b_flash}]{\begin{tikzpicture}[scale=1]
\begin{axis}[
  legend style = {at = {(0.98, 0.98)}, fill=none, fill opacity=0.9, draw opacity=1,text opacity=1, rounded corners=3pt, draw=none, font=\scriptsize,
  legend columns=2},
  legend cell align={left},
  legend image code/.code={
    \draw[#1] (0cm,0cm) -- (0.5cm,0cm);
  },
  xlabel=\textbf{training step},
  ylabel=\textbf{max. logit},
  ylabel style={yshift=-15pt},
  xlabel style={yshift=5pt},
  x tick scale label style={
    yshift=5pt
  },
  xmin=2000,
  xmax=12000,
  ymin=50,
  ymax=300,
  grid,
  grid style={dashed},
  width=.165\textwidth,
  height=.11\textwidth,
  cycle list name = color list,
  scale only axis = true,
  enlargelimits = false,
  label style={font=\scriptsize},
  tick label style={font=\scriptsize}  
  ]

\addplot [thick, black!20!blue] table {figs/data/logit_explotion_wiki40b/softmax_L12_H12_HDim64_blkSize128.dat};
\addlegendentry{n: 128}
\addplot [thick, black!20!green] table {figs/data/logit_explotion_wiki40b/softmax_L12_H12_HDim64_blkSize512.dat};
\addlegendentry{n: 512}
\addplot [thick, black!20!red] table 
{figs/data/logit_explotion_wiki40b/softmax_L12_H12_HDim64_blkSize2048_fp32.dat};
\addlegendentry{n: 2K}
\end{axis}
\end{tikzpicture}}\hfill
    \subfloat[Logit growth in MHSA for increasing $n$ in 
    LibriLight dataset.\label{fig:logit_explosion_librilight_flash}]{\begin{tikzpicture}[scale=1]
\begin{axis}[
  legend style = {at = {(0.98, 0.98)}, fill=none, fill opacity=0.9, draw opacity=1,text opacity=1, rounded corners=3pt, draw=none, font=\scriptsize,
  legend columns=2},
  legend cell align={left},
  legend image code/.code={
    \draw[#1] (0cm,0cm) -- (0.5cm,0cm);
  },
  xlabel=\textbf{training step},
  ylabel=\textbf{max. logit},
  ylabel style={yshift=-15pt},
  xlabel style={yshift=5pt},
  x tick scale label style={
    yshift=5pt
  },
  xmin=2000,
  xmax=12000,
  ymin=20,
  ymax=300,
  grid,
  grid style={dashed},
  width=.165\textwidth,
  height=.11\textwidth,
  cycle list name = color list,
  scale only axis = true,
  enlargelimits = false,
  label style={font=\scriptsize},
  tick label style={font=\scriptsize}  
  ]

\addplot [thick, black!20!blue] table {figs/data/logit_explotion_librilight/softmax_L12_H12_HDim64_blkSize128.dat};
\addlegendentry{n: 128}
\addplot [thick, black!20!green] table {figs/data/logit_explotion_librilight/softmax_L12_H12_HDim64_blkSize512.dat};
\addlegendentry{n: 512}
\addplot [thick, black!20!red] table 
{figs/data/logit_explotion_librilight/softmax_L12_H12_HDim64_blkSize2048_fp32.dat};
\addlegendentry{n: 2K}
\end{axis}
\end{tikzpicture}}\hfill
    \subfloat[Logit growth in Full-Local Attention for increasing $n$ in 
    Wiki40B dataset.\label{fig:logit_explosion_wiki40b_ls}]{\begin{tikzpicture}[scale=1]
\begin{axis}[
  legend style = {at = {(0.9, 0.98)}, fill=none, fill opacity=0.9, draw opacity=1,text opacity=1, rounded corners=3pt, draw=none, font=\scriptsize,
  legend columns=1},
  legend cell align={left},
  legend image code/.code={
    \draw[#1] (0cm,0cm) -- (0.5cm,0cm);
  },
  xlabel=\textbf{training step},
  ylabel=\textbf{max. logit},
  ylabel style={yshift=-15pt},
  xlabel style={yshift=5pt},
  x tick scale label style={
    yshift=5pt
  },
  xmin=2000,
  xmax=12000,
  ymin=50,
  ymax=300,
  grid,
  grid style={dashed},
  width=.165\textwidth,
  height=.11\textwidth,
  cycle list name = color list,
  scale only axis = true,
  enlargelimits = false,
  label style={font=\scriptsize},
  tick label style={font=\scriptsize}  
  ]

\addplot [thick, black!20!red] table {figs/data/logit_explotion_wiki40b/softmax_L12_H12_HDim64_blkSize2048_fp32.dat};
\addlegendentry{MHSA; n: 2K}
\addplot [thick, black!20!cyan] table {figs/data/logit_explotion_wiki40b/softmaxLocalAttn_L12_H12_HDim64_blkSize2048.dat};
\addlegendentry{Full-Local; n: 2K}
\addplot [thick, black!20!violet] table {figs/data/logit_explotion_wiki40b/softmaxLocalAttn_L12_H12_HDim64_blkSize8192.dat};
\addlegendentry{Full-Local; n: 8K}
\addplot [thick, black!20!brown] table 
{figs/data/logit_explotion_wiki40b/softmaxLocalAttn_L12_H12_HDim64_blkSize32768.dat};
\addlegendentry{Full-Local; n: 32K}
\end{axis}
\end{tikzpicture}} \hfill
    \subfloat[Logit growth in Full-Local Attention for increasing $n$ in 
    LibriLight dataset.\label{fig:logit_explosion_librilight_ls}]{\begin{tikzpicture}[scale=1]
\begin{axis}[
  legend style = {at = {(0.9, 1.02)}, fill=none, fill opacity=0.9, draw opacity=1,text opacity=1, rounded corners=3pt, draw=none, font=\scriptsize,
  legend columns=1},
  legend cell align={left},
  legend image code/.code={
    \draw[#1] (0cm,0cm) -- (0.5cm,0cm);
  },
  xlabel=\textbf{training step},
  ylabel=\textbf{max. logit},
  ylabel style={yshift=-15pt},
  xlabel style={yshift=5pt},
  x tick scale label style={
    yshift=5pt
  },
  xmin=2000,
  xmax=12000,
  ymin=20,
  ymax=300,
  grid,
  grid style={dashed},
  width=.165\textwidth,
  height=.11\textwidth,
  cycle list name = color list,
  scale only axis = true,
  enlargelimits = false,
  label style={font=\scriptsize},
  tick label style={font=\scriptsize}  
  ]

\addplot [thick, black!20!red] table {figs/data/logit_explotion_librilight/softmax_L12_H12_HDim64_blkSize2048_fp32.dat};
\addlegendentry{MHSA; n: 2K}
\addplot [thick, black!20!cyan] table {figs/data/logit_explotion_librilight/softmaxLocalAttn_L12_H12_HDim64_blkSize2048.dat};
\addlegendentry{Full-Local; n: 2K}
\addplot [thick, black!20!violet] table {figs/data/logit_explotion_librilight/softmaxLocalAttn_L12_H12_HDim64_blkSize8192.dat};
\addlegendentry{Full-Local; n: 8K}
\addplot [thick, black!20!brown] table 
{figs/data/logit_explotion_librilight/softmaxLocalAttn_L12_H12_HDim64_blkSize32768.dat};
\addlegendentry{Full-Local; n: 32K}
\end{axis}
\end{tikzpicture}}\\
    \subfloat[Training curves of MHSA for increasing $n$ in Wiki40B dataset.\label{fig:instability_wiki40b_flash}]{\begin{tikzpicture}[scale=1]
\begin{axis}[
  legend style = {at = {(1.03, 0.96)}, fill=none, fill opacity=0.9, draw opacity=1,text opacity=1, rounded corners=3pt, draw=none, font=\scriptsize, legend columns=2},
  legend cell align={left},
  legend image code/.code={
    \draw[#1] (0cm,0cm) -- (0.45cm,0cm);
  },
  xlabel=\textbf{training step},
  ylabel=\textbf{train loss},
  ylabel style={yshift=-15pt},
  xlabel style={yshift=5pt},
  x tick scale label style={
    yshift=5pt
  },
  xmax=12000,
  ymin=2,
  ymax=10.5,
  grid,
  grid style={dashed},
  width=.165\textwidth,
  height=.11\textwidth,
  cycle list name = color list,
  scale only axis = true,
  enlargelimits = false,
  label style={font=\scriptsize},
  tick label style={font=\scriptsize}  
  ]

\addplot [thick, black!20!yellow] table 
{figs/data/fig_instability_wiki40b/softmax_L12_H12_HDim64_blkSize512_wiki40b.dat};
\addlegendentry{n = 512}
\addplot [thick, black!20!blue] table 
{figs/data/fig_instability_wiki40b/softmax_L12_H12_HDim64_blkSize2048_wiki40b.dat};
\addlegendentry{n = 2K}
\addplot [thick, black!20!green] table 
{figs/data/fig_instability_wiki40b/softmax_L12_H12_HDim64_blkSize8192_wiki40b.dat};
\addlegendentry{n = 8K}
\addplot [thick, black!20!red] table 
{figs/data/fig_instability_wiki40b/softmax_L12_H12_HDim64_blkSize32768_wiki40b.dat};
\addlegendentry{n = 32K}
\end{axis}
\end{tikzpicture}}\hfill
    \subfloat[Training curves of MHSA for increasing $n$ in LibriLight dataset.\label{fig:instability_librilight_flash}]{\begin{tikzpicture}[scale=1]
\begin{axis}[
  legend style = {at = {(1, 0.96)}, fill=none, fill opacity=0.9, draw opacity=1,text opacity=1, rounded corners=3pt, draw=none, font=\scriptsize, legend columns=1},
  legend cell align={left},
  legend image code/.code={
    \draw[#1] (0cm,0cm) -- (0.45cm,0cm);
  },
  xlabel=\textbf{training step},
  ylabel=\textbf{train loss},
  ylabel style={yshift=-15pt},
  xlabel style={yshift=5pt},
  x tick scale label style={
    yshift=5pt
  },
  xlabel=\textbf{training step},
  ylabel=\textbf{train loss},
  xmax=20000,
  ymin=0.9,
  ymax=8,
  grid,
  grid style={dashed},
  width=.165\textwidth,
  height=.11\textwidth,
  cycle list name = color list,
  scale only axis = true,
  enlargelimits = false,
  label style={font=\scriptsize},
  tick label style={font=\scriptsize}  
  ]

\addplot [thick, black!20!blue] table 
{figs/data/fig_instability_librilight/softmax_L12_HDim64_blkSize2048.dat};
\addlegendentry{n = 2K}
\addplot [thick, black!20!green] table 
{figs/data/fig_instability_librilight/softmax_L12_HDim64_blkSize8192.dat};
\addlegendentry{n = 8K}
\addplot [thick, black!20!red] table 
{figs/data/fig_instability_librilight/softmax_L12_HDim64_blkSize32768.dat};
\addlegendentry{n = 32K}
\end{axis}
\end{tikzpicture}}\hfill
    \subfloat[Training curves of Full-Local Attention for increasing $n$ in Wiki40B dataset.\label{fig:instability_wiki40b_ls}]{\begin{tikzpicture}[scale=1]
\begin{axis}[
  legend style = {at = {(1.02, 0.96)}, fill=none, fill opacity=0.9, draw opacity=1,text opacity=1, rounded corners=3pt, draw=none, font=\scriptsize, legend columns=2},
  legend cell align={left},
  legend cell align={left},
  legend image code/.code={
    \draw[#1] (0cm,0cm) -- (0.45cm,0cm);
  },
  xlabel=\textbf{training step},
  ylabel=\textbf{train loss},
  ylabel style={yshift=-15pt},
  xlabel style={yshift=5pt},
  x tick scale label style={
    yshift=5pt
  },
  xmax=22000,
  ymin=2,
  ymax=10.5,
  grid,
  grid style={dashed},
  width=.165\textwidth,
  height=.11\textwidth,
  cycle list name = color list,
  scale only axis = true,
  enlargelimits = false,
  label style={font=\scriptsize},
  tick label style={font=\scriptsize}  
  ]

\addplot [thick, black!20!blue] table 
{figs/data/fig_instability_wiki40b/softmaxLocalAttn_L12_H12_HDim64_blkSize2048_wiki40b.dat};
\addlegendentry{n = 2K}
\addplot [thick, black!20!green] table 
{figs/data/fig_instability_wiki40b/softmaxLocalAttn_L12_H12_HDim64_blkSize8192_wiki40b.dat};
\addlegendentry{n = 8K}
\addplot [thick, black!20!red] table 
{figs/data/fig_instability_wiki40b/softmaxLocalAttn_L12_H12_HDim64_blkSize32768_wiki40b.dat};
\addlegendentry{n = 32K}
\end{axis}
\end{tikzpicture}}\hfill
    \subfloat[Training curves of Full-Local Attention for increasing $n$ in 
    LibriLight dataset.\label{fig:instability_librilight_ls}]{\begin{tikzpicture}[scale=1]
\begin{axis}[
  legend style = {at = {(1.02, 0.96)}, fill=none, fill opacity=0.9, draw opacity=1,text opacity=1, rounded corners=3pt, draw=none, font=\scriptsize, legend columns=2},
  legend cell align={left},
  legend image code/.code={
    \draw[#1] (0cm,0cm) -- (0.45cm,0cm);
  },
  ylabel style={yshift=-15pt},
  xlabel style={yshift=5pt},
  x tick scale label style={
    yshift=5pt
  },
  xlabel=\textbf{training step},
  ylabel=\textbf{train loss},
  xmax=24000,
  ymin=0.9,
  ymax=6.5,
  grid,
  grid style={dashed},
  width=.165\textwidth,
  height=.11\textwidth,
  cycle list name = color list,
  scale only axis = true,
  enlargelimits = false,
  label style={font=\scriptsize},
  tick label style={font=\scriptsize}  
  ]

\addplot [thick, black!20!blue] table 
{figs/data/fig_instability_librilight/softmaxLocalAttn_L12_HDim64_blkSize2048.dat};
\addlegendentry{n = 2K}
\addplot [thick, black!20!green] table 
{figs/data/fig_instability_librilight/softmaxLocalAttn_L12_HDim64_blkSize8192.dat};
\addlegendentry{n = 8K}
\addplot [thick, black!20!red] table 
{figs/data/fig_instability_librilight/softmaxLocalAttn_L12_HDim64_blkSize32768.dat};
\addlegendentry{n = 32K}
\end{axis}
\end{tikzpicture}}
    \caption{Logit growth and training stability at long sequence lengths on the Wiki40B (English) \citep{GuoDVA20} and LibriLight (6K split) \citep{librilight} datasets. (a,b) MHSA exhibits increasingly severe logit growth as sequence length increases. (c,d) Replacing a subset of full attention heads with local attention heads substantially reduces logit growth in Full-Local Attention. (e,f) MHSA becomes unstable at longer sequence lengths, whereas (g,h) Full-Local Attention remains stable for sequence lengths up to $n=32K$. The attention head dimension is fixed at 
    d=64 for both MHSA and Full-Local Attention. Full-Local Attention uses $11$ local 
    heads, $1$ full-attention head, and local attention span $w=100$.}
    \label{fig:instability_other_datasets}
\end{figure*}


\section{Results on Other Datasets}
\label{app:other_datasets}

To evaluate whether the observed relationship between sequence
length, attention-logit growth, and training instability 
generalizes beyond PG-19 and across modalities, we conduct 
additional experiments on the Wiki40B (English split) dataset 
\citep{GuoDVA20} and the LibriLight (6K split) speech dataset 
\citep{librilight}.

\paragraph{Datasets and Preprocessing.}
For Wiki40B, we follow the same preprocessing pipeline used 
for PG-19, normalizing the text with NMT\_NFKC and tokenizing
it using a SentencePiece unigram tokenizer with a vocabulary
size of $10K$. For LibriLight, the audio is tokenized using 
a pretrained HuBERT model \citep{HsuBTLSM21} with a 500-cluster
K-means tokenizer, resulting in a vocabulary size of $500$. In
both cases, GPT-2 Small models with the same configuration 
described in Section~\ref{sec:exp_setup} are trained 
autoregressively on the resulting token sequences.

\paragraph{Results.}
Figure~\ref{fig:instability_other_datasets} summarizes
the results. On both datasets, MHSA exhibits the same
behavior observed on PG-19. As the sequence length
increases, attention logits grow substantially
(Figures~\ref{fig:logit_explosion_wiki40b_flash} and
\ref{fig:logit_explosion_librilight_flash}), and training
eventually becomes unstable for sufficiently long
sequences (Figures~\ref{fig:instability_wiki40b_flash}
and \ref{fig:instability_librilight_flash}).

Replacing MHSA with Full-Local Attention (with the number of
local heads $H_s=11$ and the number of full heads
$H_l=1$) consistently mitigates
this behavior. Across both datasets, Full-Local Attention
substantially reduces attention-logit growth
(Figures~\ref{fig:logit_explosion_wiki40b_ls} and
\ref{fig:logit_explosion_librilight_ls}) and remains
stable at sequence lengths up to $32K$
(Figures~\ref{fig:instability_wiki40b_ls} and
\ref{fig:instability_librilight_ls}). 

These results demonstrate that the observations of 
Section~\ref{sec:exp_results} generalize across 
language and speech modeling tasks, both of which 
exhibit strong dense local dependency structure. 
This is consistent with our hypothesis: the low-rank 
bottleneck and the resulting logit explosion are 
expected to arise in any domain where tokens depend 
heavily on their immediate neighbors, and to be less 
pronounced in domains where attention patterns are 
naturally sparse or low-rank.
The consistent effectiveness of Full-Local Attention across 
language and speech settings further supports the role 
of explicit local attention as a principled remedy for 
the low-rank bottleneck in tasks with dense local 
dependency structure.

\section{Additional Evidence from Other Structured 
Attention and Positional Encoding}
\label{app:additional_evidence}

If dense local dependencies are a primary source of the
low-rank bottleneck in full self-attention, then methods
that explicitly model local interactions or bias attention
toward nearby tokens should exhibit improved training
stability. To test this prediction, we examine two classes
of existing approaches that were not designed to address
the low-rank bottleneck directly: other structured attention
mechanisms that combine local and full attention, and
positional encodings that encourage attention to nearby
tokens.

Figure~\ref{fig:instability_struct_attn} compares the
training stability of structured attention mechanisms
inspired by \citet{GuoQXZ19} and \citet{abs-2004-05150}.
Although these architectures were originally proposed for
data-efficient training or efficient long-context modeling, 
they share a common
characteristic with Full-Local Attention: a portion of the
attention capacity is dedicated to local interactions.
Consistent with our hypothesis, both approaches exhibit
substantially improved stability compared to standard
MHSA during long-sequence training.

Figure~\ref{fig:instability_pos_encoding} presents a
similar comparison for positional encoding schemes.
Models using RoPE~\citep{SuALPBL24} and ALiBi~\citep{PressSL22} 
remain considerably more
stable than a model without positional encoding (NoPE).
Unlike Full-Local Attention or structured attention, these
methods do not explicitly introduce local attention
heads. However, both RoPE and ALiBi bias the attention
mechanism toward nearby tokens, effectively increasing
the emphasis placed on local dependencies. The improved
stability observed with these encodings is therefore
consistent with the view that local interactions play a
central role in the instability of long-sequence
training.

Taken together, these results provide complementary
evidence for our central hypothesis. Across a diverse set
of architectures and positional encoding schemes,
approaches that explicitly model or emphasize local
dependencies consistently exhibit improved training
stability. This suggests that explicitly modeling or emphasizing 
local dependencies is a broadly effective principle for 
mitigating the low-rank bottleneck and improving 
long-sequence training stability, regardless of the 
specific architectural mechanism employed.

\begin{figure}[t!]
    \centering
    \subfloat[Structured attention mechanisms that explicitly
    combine local and full attention improve training stability. \label{fig:instability_struct_attn}]{\begin{tikzpicture}[scale=1]
\begin{axis}[
  legend style = {at = {(1.02, 1)}, fill=none, fill opacity=0.9, draw opacity=1,text opacity=1, rounded corners=3pt, draw=none, font=\scriptsize},
  legend cell align={left},
  legend image code/.code={
    \draw[#1] (0cm,0cm) -- (0.5cm,0cm);
  },
  xlabel=\textbf{training step},
  ylabel=\textbf{train loss},
  ylabel style={yshift=-15pt},
  xlabel style={yshift=5pt},
  x tick scale label style={
    yshift=5pt
  },
  xmax=14000,
  ymax=12,
  ymin=2,
  ytick={2, 6, 10},
  grid,
  grid style={dashed},
  width=.17\textwidth,
  height=.11\textwidth,
  cycle list name = color list,
  scale only axis = true,
  enlargelimits = false,
  label style={font=\scriptsize},
  tick label style={font=\scriptsize}  
  ]

\addplot [thick, black!20!blue] table 
{figs/data/fig_structured_attn/softmax_L12_H12_HDim64_blkSize2048.dat};
\addlegendentry{MHSA}
\addplot [thick, black!20!red] table 
{figs/data/fig_structured_attn/softmaxConstrainedAttn_L12_H12_HDim64_blkSize2048_attnSpan100.dat};
\addlegendentry{Guo et al. 2019}
\addplot [thick, black!20!green] table 
{figs/data/fig_structured_attn/softmaxLocalFullInt_L12_H12_HDim64_I3_blkSize2K_attnSpan100.dat};
\addlegendentry{Beltagy et al. 2020}
\end{axis}
\end{tikzpicture}}\hfill
    \subfloat[Positional encodings that bias attention toward
    nearby tokens improve training stability. \label{fig:instability_pos_encoding}]{\begin{tikzpicture}[scale=1]
\begin{axis}[
  legend style = {at = {(1.02, 1)}, fill=none, fill opacity=0.9, draw opacity=1,text opacity=1, rounded corners=3pt, draw=none, font=\scriptsize},
  legend cell align={left},
  legend image code/.code={
    \draw[#1] (0cm,0cm) -- (0.5cm,0cm);
  },
  xlabel=\textbf{training step},
  ylabel=\textbf{train loss},
  ylabel style={yshift=-15pt},
  xlabel style={yshift=5pt},
  x tick scale label style={
    yshift=5pt
  },
  xmax=14000,
  ymin=2,
  ymax=12,
  ytick={2, 6, 10},
  grid,
  grid style={dashed},
  width=.17\textwidth,
  height=.11\textwidth,
  cycle list name = color list,
  scale only axis = true,
  enlargelimits = false,
  label style={font=\scriptsize},
  tick label style={font=\scriptsize}  
  ]

\addplot [thick, black!20!blue] table 
{figs/data/fig_instability_otherPosEncoding/softmaxNoPE_H12_HDim64_blkSize8K.dat};
\addlegendentry{NoPE}
\addplot [thick, black!20!red] table 
{figs/data/fig_instability_otherPosEncoding/softmaxRoPE_H12_HDim64_blkSize8K_base40K.dat};
\addlegendentry{RoPE}
\addplot [thick, black!20!green] table 
{figs/data/fig_instability_otherPosEncoding/softmaxALiBi_H12_HDim64_blkSize8K.dat};
\addlegendentry{ALiBi}
\end{axis}
\end{tikzpicture}}
    \caption{Additional evidence for the role of local dependencies
    in long-sequence training stability. (a) Structured attention 
    mechanisms that allocate dedicated capacity to local interactions, 
    such as those of \citet{GuoQXZ19} and \citet{abs-2004-05150},
    exhibit improved training stability. (b) Positional encodings
    such as RoPE and ALiBi, which bias attention toward nearby
    tokens, are substantially more stable than NoPE (no positional 
    encoding). These observations are consistent with the hypothesis
    that dense local dependencies are a primary cause of 
    attention-logit growth and that explicitly modeling or 
    emphasizing local dependencies alleviates the low-rank 
    bottleneck, thereby reducing logit explosion and improving 
    long-sequence training stability.}
    \label{fig:instability_other_scheme}
\end{figure}

\begin{figure}[h!]
    \centering
    \begin{tikzpicture}[scale=1]
\begin{axis}[
  legend style = {at = {(0.97, 0.95)}, fill=none, fill opacity=0.9, draw opacity=1,text opacity=1, rounded corners=3pt, draw=none, font=\scriptsize},
  legend cell align={left},
  y filter/.code={\pgfmathparse{#1}\pgfmathresult},
  xlabel=\textbf{training step},
  ylabel=\textbf{train loss},
  ylabel style={yshift=-15pt},
  xlabel style={yshift=5pt},
  x tick scale label style={
    yshift=5pt
  },
  xmax=25000,
  ymin=2.5,
  ymax=7,
  grid,
  grid style={dashed},
  width=.22\textwidth,
  height=.15\textwidth,
  cycle list name = color list,
  scale only axis = true,
  enlargelimits = false,
  label style={font=\scriptsize},
  tick label style={font=\scriptsize}  
  ]

\addplot [thick, black!20!blue] table 
{figs/data/fig_alternative_stabilization/softmaxLocalAttn_L12_H12_HDim64_attnSpan100_blkSize2048.dat};
\addlegendentry{FL-attn}
\addplot [thick, black!20!green] table 
{figs/data/fig_alternative_stabilization/softmaxQKnorm_H12_HDim64_blkSize2048.dat};
\addlegendentry{MHSA + QK-norm}
\addplot [thick, black!20!red] table 
{figs/data/fig_alternative_stabilization/softmaxZloss_H12_HDim64_blkSize2048.dat};
\addlegendentry{MHSA + Z-loss}
\addplot [thick, black!20!cyan] table 
{figs/data/fig_alternative_stabilization/softmaxAdaGC_H12_HDim64_blkSize2048.dat};
\addlegendentry{MHSA + AdaGC}
\end{axis}
\end{tikzpicture}
    \caption{Evaluation of existing training stabilization 
    methods for long-sequence transformer training. The 
    evaluated methods include (1) MHSA with QK-normalization,
    (2) MHSA with Z-loss, and (3) MHSA with the AdaGC 
    optimizer.}
    \label{fig:comp_alternative}
\end{figure}

\section{Existing Training Stabilization Methods Do Not 
Address the Root Cause}
\label{app:alt_train_stab}

Several methods have been proposed to improve the
stability of transformer training, including
QK-normalization \citep{HenryDPC20}, Z-loss
\citep{ChowdheryNDBMRBCSGSSTMRBTSPRDHPBAI23}, and the
AdaGC optimizer \citep{abs-2502-11034}. However, these
methods primarily target the symptoms of instability,
such as large attention logits or unstable optimization,
rather than the underlying representational bottleneck
identified in this work.

To evaluate their effectiveness for long-sequence
training, we trained the baseline transformer on the
PG-19 dataset using each of these techniques.
Figure~\ref{fig:comp_alternative} shows the resulting
training curves. Z-loss and AdaGC fail to prevent
long-sequence training instability. QK-normalization successfully
stabilizes training, but convergence is substantially
slower than with Full-Local Attention. After $25K$ training
steps, QK-normalization reaches a training loss of
approximately $3.16$ (perplexity $23.57$), whereas
Full-Local Attention achieves a substantially lower loss of
approximately $2.77$ (perplexity $15.96$), corresponding
to roughly a $32\%$ reduction in perplexity.

These results suggest that suppressing the symptoms 
of logit explosion is less effective than addressing 
its underlying cause. In contrast to existing
stabilization techniques, adding explicit local
attention heads directly alleviates the low-rank 
bottleneck by allocating dedicated attention capacity 
to dense local dependencies, reducing the need for 
extreme logit magnitudes without sacrificing 
convergence speed. These results indicate that 
addressing the root cause of logit explosion is both 
more effective and more efficient than attempting to 
compensate for its downstream consequences.

\end{document}